\documentclass[10pt,twocolumn,letterpaper]{article}

\usepackage{iccv}
\usepackage{times}
\usepackage{epsfig}
\usepackage{graphicx}
\usepackage{amsmath}
\usepackage{amssymb}

\usepackage{graphicx}
\usepackage{subcaption}
\usepackage{float}
\usepackage{lscape}                                         % Useful for wide tables or figures.

\usepackage[lined,ruled,linesnumbered]{algorithm2e}

\usepackage{booktabs}                   % Publication quality tables
\usepackage{multirow}

\usepackage{paralist}
\usepackage{enumitem}

\usepackage{bm}                          % Make bold, italic math symbols
\usepackage{epsfig}                      % for figures
\usepackage{graphicx}                  % another package that works for figures
\usepackage{times}
\usepackage{mathtools}
\usepackage{amssymb,amsmath}   % Short math guide for LaTeX ftp://ftp.ams.org/pub/tex/doc/amsmath/short-math-guide.pdf

\usepackage{units}
\usepackage{color}

\usepackage{comment}

\usepackage{url}  % Hyphenation of URLs.

\usepackage{xspace}
\usepackage[table]{xcolor}
\usepackage{setspace}

\usepackage[pagebackref=true,breaklinks=true,colorlinks,bookmarks=false,citecolor=blue,linkcolor=blue]{hyperref}

\newlength\paramarginsize
\newlength\figmarginsize
\newlength\secmarginsize
\newlength\figcapmarginsize
\newlength\tabmarginsize

\newcommand{\figmargin}{\vspace{\figmarginsize}}

\newcommand{\tabmargin}{\vspace{\tabmarginsize}}

\newcommand {\first}[1]{{\color{red}\textbf{#1}}}
\newcommand {\second}[1]{{\color{blue}\underline{#1}}}

\newcommand{\heading}[1]
{
\vspace{1mm}\noindent\textbf{#1}
}

\newcommand{\secref}[1]{Sec.~\ref{Sec:#1}}
\newcommand{\figref}[1]{Fig.~\ref{fig:#1}} 
\newcommand{\tabref}[1]{Table~\ref{tab:#1}}
\newcommand{\eqnref}[1]{\eqref{Eq:#1}}

\long\def\ignorethis#1{}

\definecolor{aqua}{rgb}{0.1, 0.5, 0.7}
\definecolor{tealblue}{rgb}{0.212, 0.459, 0.533}
\definecolor{purple}{rgb}{0.5, 0, 0.5}

\def\xi{\mathbf{x}_i}

\def\cost{L}
\def\weight{w}
\def\coststable{\cost_{C^0}}
\def\costsmooth{\cost_{C^1}}
\def\costprotrusion{\cost_{p}}
\def\costdistortion{\cost_{d}}
\def\costflow{\cost_{f}}
\def\weightstable{\weight_{C^0}}
\def\weightsmooth{\weight_{C^1}}
\def\weightprotrusion{\weight_{p}}
\def\weightdistortion{\weight_{d}}
\def\weightflow{\weight_{f}}
\newcommand{\flow}[2]{F^{#2}_{#1}}
\newcommand{\flowtilde}[2]{\tilde{F}^{#2}_{#1}}

\graphicspath{{figure}, {images}, {example}}

\iccvfinalcopy % *** Uncomment this line for the final submission

\def\iccvPaperID{5887} % *** Enter the ICCV Paper ID here

\ificcvfinal\fi

\begin{document}

%%%%%%%%% TITLE
\title{Deep Online Fused Video Stabilization}

\author{
Zhenmei Shi\\
University of \\Wisconsin Madison\\
\and
Fuhao Shi\\
Google\\
\and
Wei-Sheng Lai\\
Google\\
\and
Chia-Kai Liang\\
Google\\
\and
Yingyu Liang\\
University of \\Wisconsin Madison\\
}

\maketitle

% Remove page # from the first page of camera-ready.
\ificcvfinal\thispagestyle{empty}\fi

%%%%%%%%% ABSTRACT
\begin{abstract}
   We present a deep neural network (DNN) that uses both sensor data (gyroscope) and image content (optical flow) to stabilize videos through unsupervised learning. The network fuses optical flow with real/virtual camera pose histories into a joint motion representation. Next, the LSTM cell infers the new virtual camera pose, which is used to generate a warping grid that stabilizes the video frames.
   We adopt a relative motion representation as well as a multi-stage training strategy to optimize our model without any supervision. 
   To the best of our knowledge, this is the first DNN solution that adopts both sensor data and image content for video stabilization. 
   We validate the proposed framework through ablation studies and demonstrate that the proposed method outperforms the state-of-art alternative solutions via quantitative evaluations and a user study. 
   Check out our video results and dataset at our \href{https://zhmeishi.github.io/dvs/}{website}.
\end{abstract}

%%%%%%%%% BODY TEXT
\section{Introduction}
Videos captured with a hand-held device are often shaky. With the growing popularity of casual video recording, live streaming, and movie-making on hand-held smartphones, effective and efficient video stabilization is crucial for improving overall video quality.

However, high-quality stabilization remains challenging due to complex camera motions and scene variations.
The existing video stabilization systems can be generally categorized into image-based and sensor-based methods. 
Image-based methods output a smooth camera path by extracting camera motions from sparse image features~\cite{Grundmann-CVPR-2011,Liu-TOG-2009,Liu-TOG-2013} or dense optical flow~\cite{choi2020deep,wang2018deep,yu2019robust,yu2020learning,zhao2020pwstablenet}.
These methods offer nonlinear flexibility on motion compensations.
However, they often fail when there are complex motions such as parallax, or no reliable features in the scene, and produce visible non-rigid  distortions and artifacts due to lack of reliable rigidity constraints.
The sensor-based methods use motion sensor data, e.g., gyroscope and accelerometer, to obtain accurate motion.
They are free from scene contents and can achieve impressive stabilization results with effective distortion corrections~\cite{fusedEIS, Thivent18}.
However, these methods deliver homographies to stabilize the plane at infinity and do not adapt to the scene depth, leading to more residual parallax motions for close scenes.

\begin{figure*}
    \centering
    \footnotesize
    \begin{tabular}{cc}
        \includegraphics[width=0.43\linewidth]{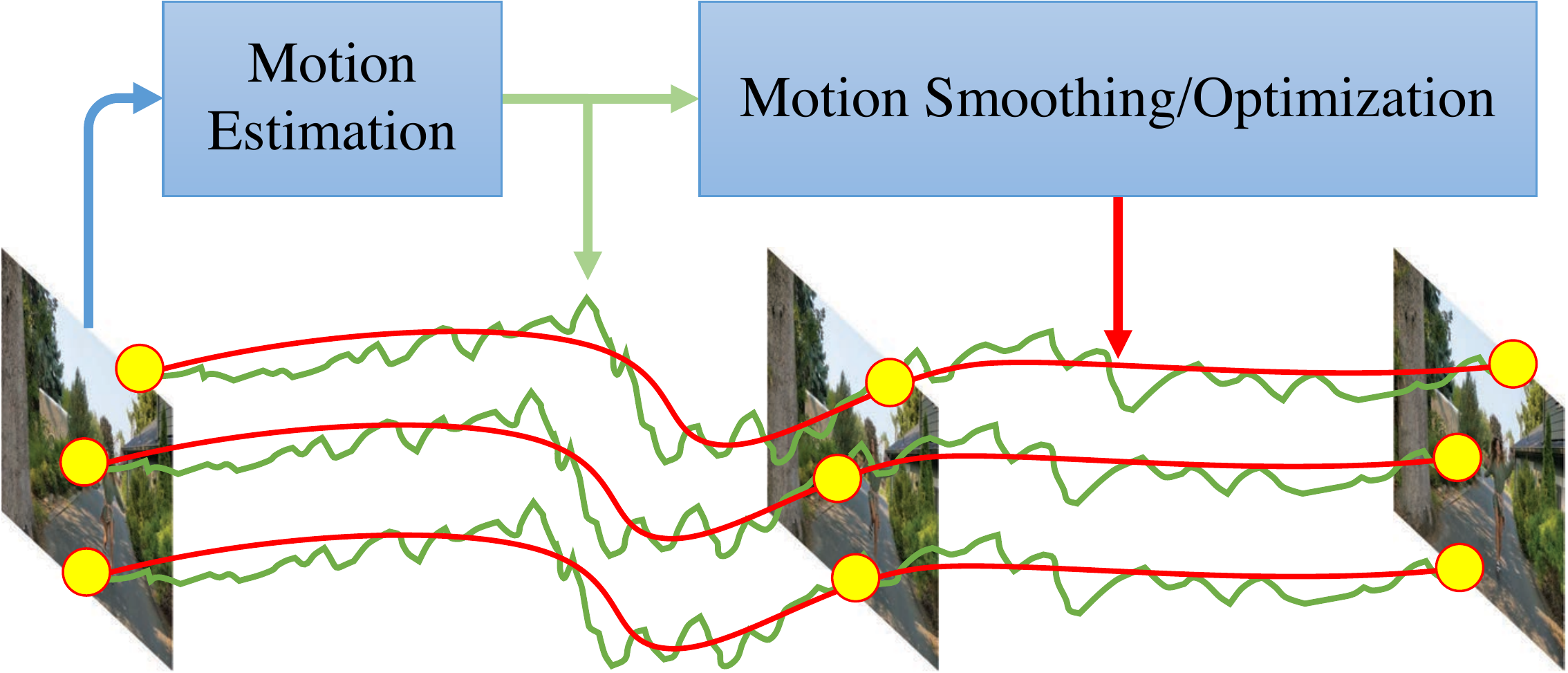} & 
        \includegraphics[width=0.43\linewidth]{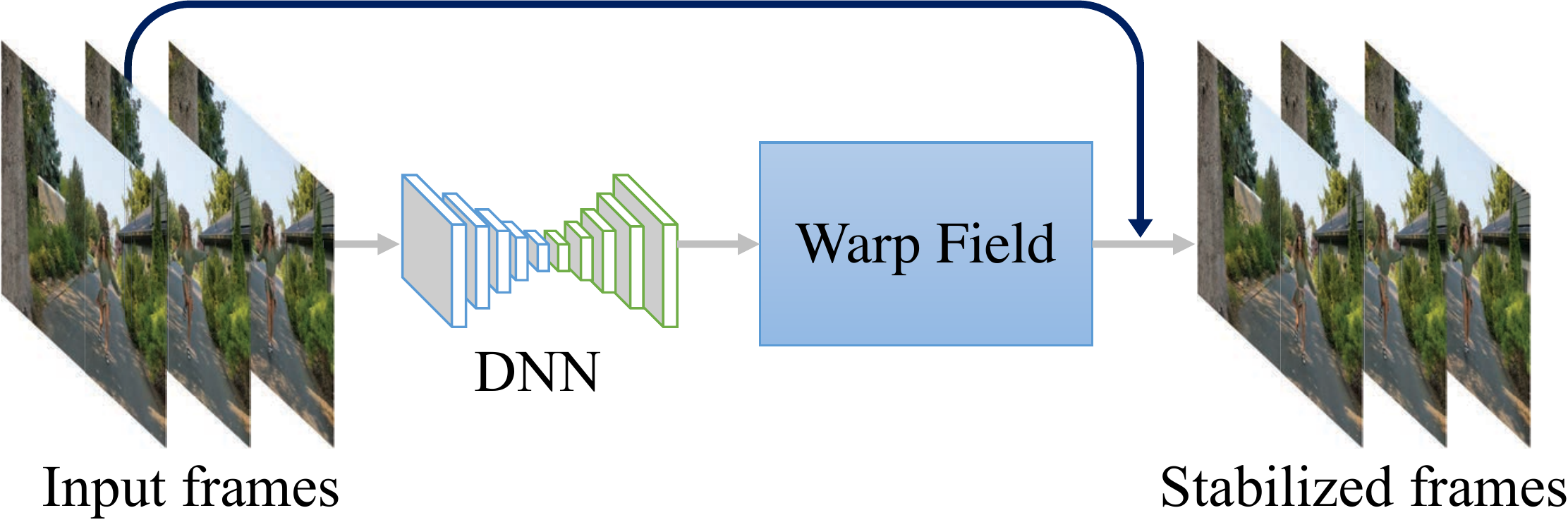} 
        \\
        (a) Conventional methods &
        (b) Learning-based methods 
        \\
        \multicolumn{2}{c}{\includegraphics[width=0.9\linewidth]{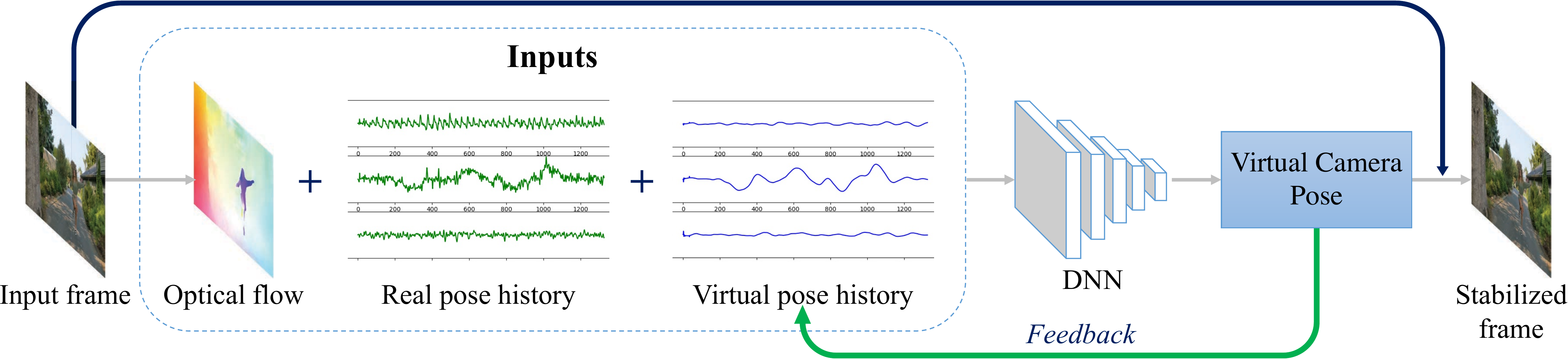}} \\
        \multicolumn{2}{c}{(c) Deep-FVS (ours)}
    \end{tabular}
    \figmargin
    \caption{
        \textbf{Comparisons of existing video stabilization methods and the proposed method.}
        (a) Conventional video stabilization methods~\cite{Grundmann-CVPR-2011,Liu-TOG-2013} estimate camera motions based on image feature trajectories and find a smooth camera path to render a stabilized video.
        (b) Learning-based approaches~\cite{choi2020deep,wang2018deep,xu2018deep,yu2020learning,zhao2020pwstablenet} learn deep networks to predict warp fields for warping the input video.
        (c) The proposed Deep-FVS learns to stabilize a video by fusing the optical flow and gyroscope data. 
    }
    \label{fig:overview}
    \figmargin
\end{figure*}

In this work, we present an efficient deep Fused Video Stabilization (deep-FVS) framework to fuse the two motion sources (image content and motion sensor) and draw benefits from both ends.
On one hand, the network outputs a single virtual camera pose instead of dense warping flow, and the videos are then stabilized by warping the sensor-based real camera motions towards this virtual pose.
In this way, the motion rigidity is naturally preserved and rolling shutter distortion is corrected without artifacts (e.g., wobbling).
On the other hand, the network learns to jointly minimize both camera pose smoothness and optical flow losses.
Thus, it automatically adjusts to different scenes (e.g., depth variations) to minimize the residual motions.
Our network is trained with unsupervised learning with carefully designed loss functions and a multi-stage training procedure.
\figref{overview} shows an overview of conventional methods~\cite{Grundmann-CVPR-2011,Liu-TOG-2013}, recent learning-based approaches~\cite{choi2020deep,wang2018deep,xu2018deep,yu2020learning,zhao2020pwstablenet}, and the proposed deep-FVS.

As the existing datasets~\cite{Liu-TOG-2013, wang2018deep} do not record the sensor data, we collect a new video dataset that contains videos with both gyroscope and OIS data for training and evaluation.
Our dataset covers diverse scenarios with different illumination conditions and camera/subject motions.
The sensor data and video frames are accurately aligned through calibration.
We evaluate the proposed solution objectively and subjectively and show that it outperforms state-of-the-art methods by generating more stable and distortion-free results.

This paper makes the following contributions:
\begin{compactitem}
\item The first DNN-based framework that fuses motion sensor data and optical flow for online video stabilization.
\item An unsupervised learning process with multi-stage training and relative motion representation.
\item A benchmark dataset that contains videos with gyroscope and OIS sensor data and covers various scenarios. Both the dataset and code will be publicly released.
\end{compactitem}
\section{Related Work}
\heading{Conventional methods.}
Classical video stabilization algorithms typically involve motion estimation, camera path smoothing, and video frame warping/rendering steps~\cite{Morimoto1998}. Some solutions also  correct the rolling shutter distortions \cite{Grundmann2012calibration,Karpenko-CSTR-2011,Liang2008analysis}.
Those methods can be categorized into 3D, 2D, and 2.5D approaches based on motion estimation.

The 3D approaches model the camera poses and estimate a smooth virtual camera trajectory in the 3D space.
To find 6DoF camera poses, several techniques have been adopted, including projective 3D reconstruction~\cite{Buehler-CVPR-2001}, depth camera~\cite{liu2012video}, structure from motion~\cite{Liu-TOG-2009}, and light-field~\cite{smith2009light}.
While 3D approaches can handle parallax and produce high-quality results, they often entail expensive computational costs or require specific hardware devices.

The 2D approaches represent and estimate camera motions as a series of 2D affine or perspective transformations \cite{ Grundmann-CVPR-2011, Liu-TOG-2013,Matsushita-PAMI-2006}.
Robust feature tracking and outlier rejection are applied to obtain reliable  estimation~\cite{zhang2016robust}. Liu et al.~\cite{liu2014steadyflow} replace feature trajectories with optical flows to handle spatially-variant motion.
Early approaches apply low-pass filters to smooth individual motion parameters~\cite{chang2006robust, Matsushita-PAMI-2006}, while recent ones adopt $\mathcal{L}_1$ optimization~\cite{Grundmann-CVPR-2011} and joint optimization with bundled local camera paths~\cite{Liu-TOG-2013}.
Some hybrid 2D-3D approaches exploit the subspace constraints~\cite{Liu-TOG-2011} and epipolar geometry~\cite{goldstein2012video}.
Zhuang et al.~\cite{zhuang20195d} smooth 3D rotation from the gyroscope and stabilize the residual 2D motion based on feature matching.

The above methods often process a video offline, which are not suitable for live-streaming and mobile use cases.
Liu et al.~\cite{liu2016meshflow} propose a MeshFlow motion model with only one frame latency for online video stabilization.
A mobile online solution using both the OIS and EIS is developed in~\cite{fusedEIS}.
In this work, we utilize the OIS, gyroscope, and optical flow to learn a deep network for stabilization.
Our online method has only 10 frames latency and does not require per-video optimization.

\begin{figure*}[t]
    \centering
    \includegraphics[width=0.95\textwidth]{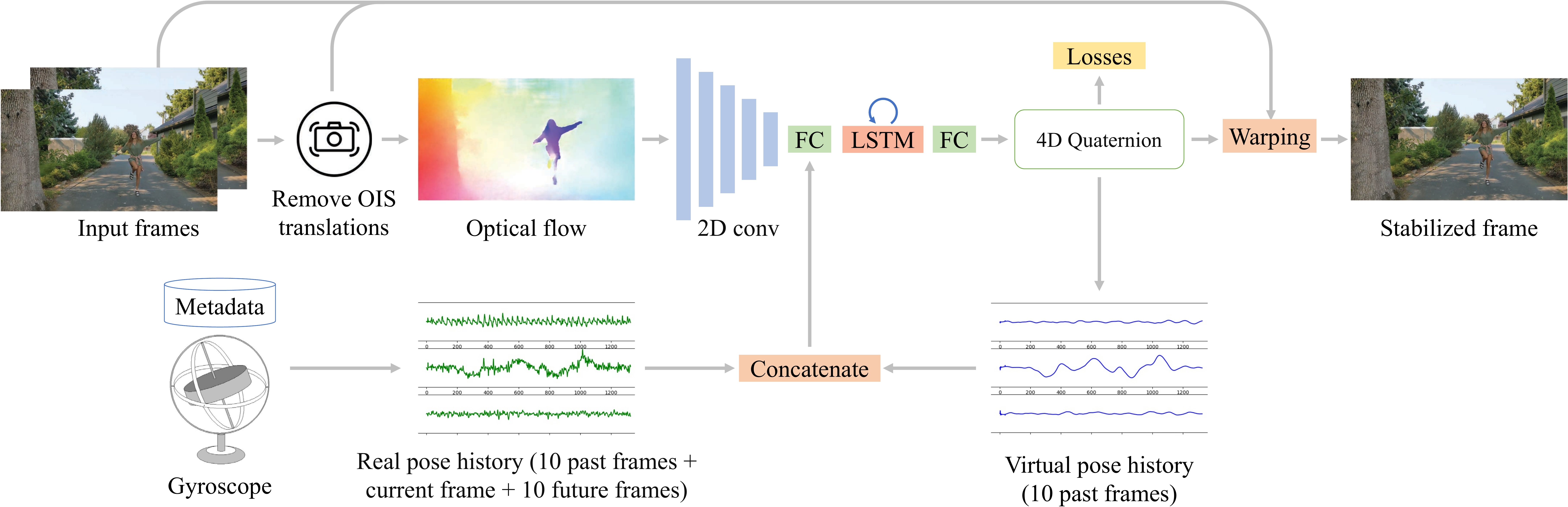}
    \figmargin
    \caption{\textbf{Overview of deep-FVS.}
    Given an input video, we first remove the OIS translation to extract the raw optical flow. We also obtain the real camera poses from the gyroscope and convert it to a relative quaternion.
    An encoder with 2D convolutions embeds optical flows to a latent representation, which is then concatenated with the real and virtual camera poses.
    This joint motion representation is fed to a LSTM cell and FC layers to predict the new virtual camera pose as a quaternion.
    Finally, we warp the input frame based on the OIS and virtual camera pose to generate the stabilized frame.
    }
    \label{fig:system}
    \figmargin
\end{figure*}

\heading{Learning-based methods.}
With the success of deep learning on image recognition~\cite{he2016deep, long2015fully, ren2015faster}, DNNs have been adopted to several computer vision tasks and achieved state-of-the-art performance.
However, DNN based video stabilization still does not attract much attention, mainly due to the lack of proper training data.
Wang et al.~\cite{wang2018deep} collect the DeepStab dataset with 60 pairs of stable/unstable videos, and train a deep CNN to predict mesh-grids for warping the video.
Instead of predicting low-resolution mesh-grids, the PWStableNet~\cite{zhao2020pwstablenet} learns dense 2D warping fields to stabilize the video.
Xu et al.~\cite{xu2018deep} train a generative adversarial network to generate a steady frame as guidance and use the spatial transformer network to extract the affine transform for warping the video frames.
Yu and Ramamoorthi~\cite{yu2019robust} take optical flows as input and optimize the weights of a deep network to generate a warp field for each specific video.
They further train a stabilization network that can be generalized test videos without optimization~\cite{yu2020learning}. 
Choi et al.~\cite{choi2020deep} learn a frame interpolation model to iteratively interpolate the input video into a stable one without cropping.

These learning-based methods learn to stabilize videos from the video content and optical flow. Their performance heavily depends on the training data and can suffer from visible distortion for large motions (e.g., running).
In contrast, we use the gyroscope to compensate camera motions and utilize optical flow to correct the residual motions from scene geometry.
\section{Deep Fused Video Stabilization}\label{Sec:DeepFVS}
The overview of our method is shown in~\figref{system}.
We first process the gyroscope and OIS reading so that we can query the real camera extrinsic (i.e., rotation) and intrinsic (i.e., principal point offsets) at arbitrary timestamps (\secref{SensorProcessing}).
We then remove the OIS translations on the input video and extract optical flows from the raw video frames (\secref{OIS-freeOpticalFlow}).
The optical flows are encoded to a latent space via 2D convolutional layers and concatenated with the real camera poses within a temporal window and the previous virtual camera poses as a joint motion representation (\secref{Representation} and~\ref{Sec:RelativeRotation}).
Next, we feed this joint motion representation to an LSTM cell and a few fully-connected layers to predict a virtual camera pose at the current timestamp.
Finally, we use a grid-based warping method similar to Karpenko et al.~\cite{Karpenko-CSTR-2011} to warp the input frame to the stabilized rolling shutter corrected domain using the input camera rotations, OIS movement, and the predicted virtual camera poses (supplementary material Section 3).
Our solution stabilizes a video frame-by-frame and is suitable for online processing.

During the training stage, we randomly select long sub-sequences from the training videos, and optimize our DNN with a set of loss functions without any ground-truth videos or camera poses for supervision (\secref{LossFuction}).
To stabilize the training, we adopt a multi-stage training strategy to constrain the solution space (\secref{MutliStageTraining}).

\subsection{Gyroscope and OIS Pre-processing} \label{Sec:SensorProcessing}
In our dataset, the gyroscope ($\omega_x, \omega_y, \omega_z, t$) and OIS ($o_x, o_y, t$) are sampled at 200 Hz, where $\omega$ is the angular velocity, and $o_x, o_y$ are the OIS movements.
The camera rotation is integrated by $R(t) = S\omega(t) * R(t-S)$, where $S$ is the sampling interval (5ms). 
We represent the rotation as a 4D quaternion and save it in a queue. 
To obtain the camera rotation at an arbitrary timestamp $t_f$, we first locate the two consecutive gyro samples $a, b$ in the queue such that $t_a \leq t_f \leq t_b$, and obtain $R(t_f)$ by applying a spherical linear interpolation (SLERP): 
\begin{equation}
R(t_f) = \mathtt{SLERP}(R(t_a), R(t_b), (t_b - t_f)/(t_b - t_a)).
\end{equation}
Similarly, $O(t)$ is calculated from a linear interpolation between $O(t_a)$ and $O(t_b)$.

\subsection{Camera Pose Representation} \label{Sec:Representation}
We represent a camera pose as $P = (R, O)$, where $R$ is the camera rotation and $O=(o_x, o_y)$ is a 2D offset to the camera principal point $(u,v)$. Given a 3D world coordinate $X$, the projected point on the 2D image at timestamp $t$ is
\begin{align}
    x = K(t)R(t)X,
\end{align}
where $K(t)=[f, 0, u + o_x(t);0, f, v + o_y(t); 0, 0, 1]$ is the intrinsic matrix with focal length $f$. 

Given a real camera pose $P_r=(R_r, O_r)$ and virtual one $P_v = (R_v, O_v)$, the transformation of a point from the real camera space to the virtual (stabilized) one is
\begin{equation} \label{Eq:projection}
x_v = K_v(t)R_v(t)R_r^{-1}(t)K_r^{-1}(t)x_r,
\end{equation}
where $x_r, x_v$ are the 2D image coordinates at real and virtual camera spaces, respectively. 
In all the experiments, we normalize $f=1511.8$ for both the real and virtual cameras.

\section{Unsupervised Learning for Sensor Fusion} \label{Sec:ModelLensing}
This section introduces the core of our deep fused video stabilization network.
As shown in~\figref{system}, our network consists of a sequence of 2D convolutional layers to encode the optical flow, an LSTM cell to fuse the latent motion representation and maintain temporal information, and fully-connected layers to decode the latent representation to virtual camera poses.
The detailed network configuration is provided in the supplementary material.

We first extract the OIS-free optical flow from the input frames and OIS data (\secref{OIS-freeOpticalFlow}) and map it to a low-dimensional representation $z$. Meanwhile, we extract the past and future real camera rotation history $H_r$ and the past virtual rotation history $H_v$ from the queues (\secref{RelativeRotation}).
We define the joint motion representation as $[z, H_r, H_v]$
and feed it into the LSTM to predict an incremental rotation $\Delta R_v(t)$ to the previous virtual pose $R_v(t-\Delta t)$, where $\Delta t$ is fixed to 40ms in our experiments and is invariant to the video frame rate.
Note we set the virtual offset $O_v$ to 0. 
The final virtual pose is then calculated as $P_v = (\Delta R_v(t) R_v(t- \Delta t), O_v)$ and used to generate the warping grid.
It is also pushed into the virtual pose queue as the input for later frames. 
We can interpret the LSTM, virtual pose prediction, and frame warping steps as a decoder that maps the current motion state $[z, H_r, H_v]$ to a stabilized frame.

\subsection{OIS-free Optical Flow} \label{Sec:OIS-freeOpticalFlow}
Camera motions in the input videos are compensated by the OIS to reduce the motion blur.
Although the OIS movement depends on the hand motion, the offset $O_r$ is different at each scanline due to the rolling shutter and more like a random noise (see the supplementary materials for more discussions). 
It is non-trivial to let the network learn to associate the local offset with the principal point changes.

To address this issue, we remove OIS motions when estimating the optical flow such that the input to our model contains only the camera and object motions.
Specifically, we denote the position of a pixel in frame $n$ as $x_{r,n}$ and its corresponding pixel in frame $n+1$ as $y_{r,n+1}$.
The raw forward optical flow can be represented as 
\begin{align}
    \flowtilde{n}{n+1} = y_{r,n+1} - x_{r,n}.
\end{align}
By reverting the OIS movement at the pixel's timestamp (which depends on the y-coordinate due to the rolling shutter readout), $x_{r,n}$ and $y_{r,n+1}$ are mapped to $x_{r,n} - O(t_{x_{r,n}})$ and $y_{r,n+1} - O(t_{y_{r,n+1}})$, respectively.
The forward optical flow is then adjusted to 
\begin{align}
    \flow{n}{n+1} 
    &= (y_{r,n+1} - O(t_{y_{r,n+1}})) - (x_{r,n} - O(t_{x_{r,n}})) \nonumber \\
    &= \flowtilde{n}{n+1} - (O(t_{y_{r,n+1}}) - O(t_{x_{r,n}})).
\end{align}
The backward flow is adjusted similarly.
We use the FlowNet2~\cite{ilg2017flownet} to extract optical flows in our experiments.

\subsection{Relative Rotation based Motion History} \label{Sec:RelativeRotation}
To obtain the real and virtual pose histories $[H_r, H_v]$ at a timestamp $t$, we first sample $N$ past and future timestamps from the gyro queue (\secref{SensorProcessing}) and obtain the real absolute camera rotations $H_{r, \text{absolute}} = (R_r(t-N\Delta t),...,R_r(t),..., R_r(t+N\Delta t))$. 
Meanwhile, we sample the virtual pose queue to obtain the virtual camera pose history as $H_{v, \text{absolute}} = (R_v(t-N\Delta t),...,R_v(t-\Delta t))$.

One key novelty here is to convert the absolute poses, which are integrated from the very first frame, into a \emph{relative} rotation w.r.t. the current real camera pose:
\begin{align}
    H_r &= H_{r, \text{absolute}} * R_r^{-1}(t), \\
    H_v &= H_{v, \text{absolute}} * R_r^{-1}(t).
\end{align}
The network output is also a relative rotation to the previous virtual camera pose. 
Therefore, our model only needs to learn the first order pose changes and is invariant to the absolute poses. 
Our experiments show that this relative rotation representation leads to more stable predictions and provides a much better generalization (\secref{AblationStudy}).  

\begin{figure}
    \centering
    \includegraphics[width=\linewidth]{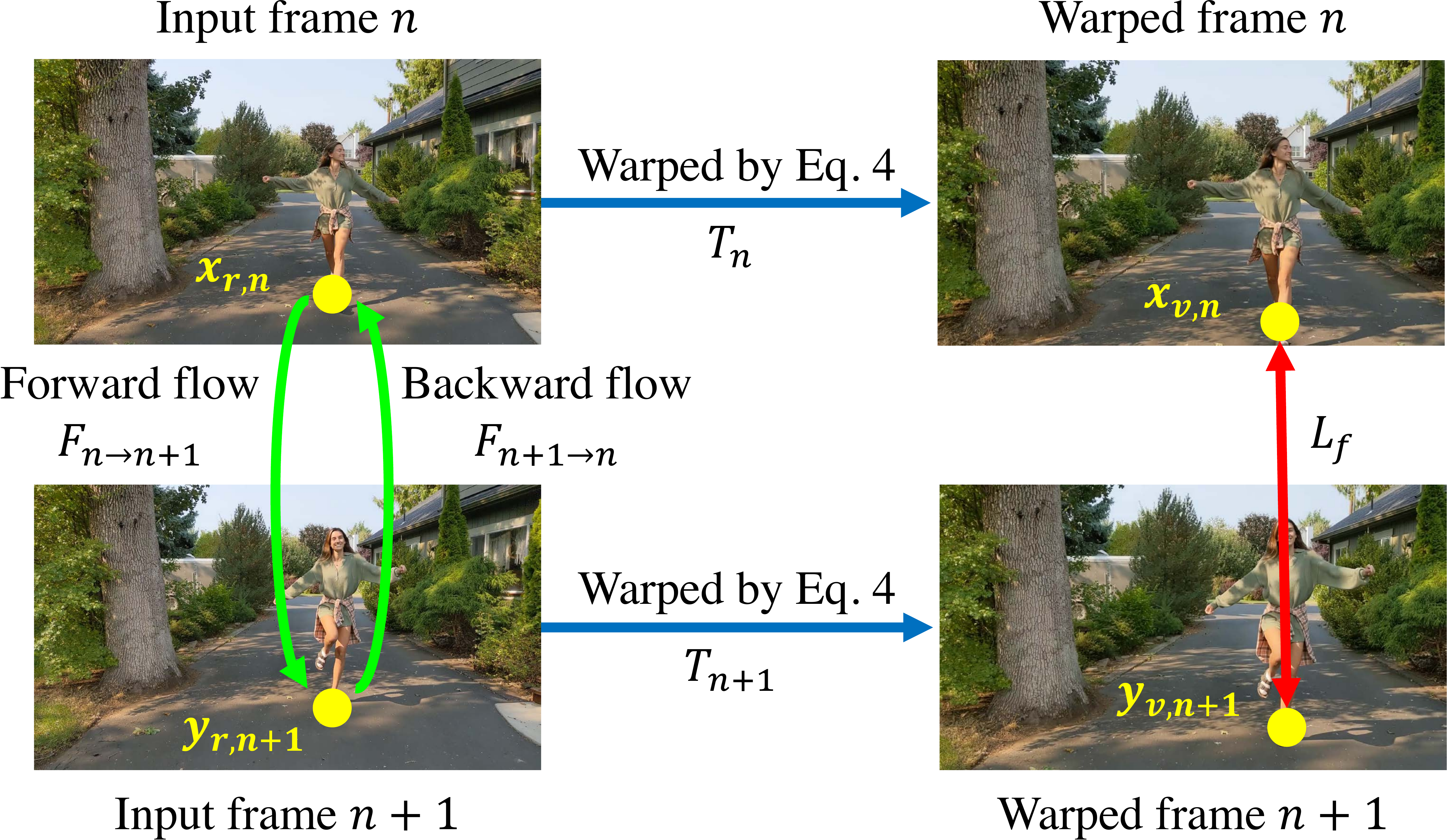}
    \figmargin
    \vspace{-3mm}
    \caption{\textbf{Optical flow loss.} The optical flow loss aims to minimize the distance between $x_{v,n}$ and $y_{v,n+1}$ in the virtual camera space. By incorporating the forward and backward flows, we define our optical flow loss as in~\eqnref{flow_loss}. }
    \label{fig:flow_loss}
    \figmargin
\end{figure}

\subsection{Loss Functions} \label{Sec:LossFuction}
We define the following loss functions to train our network. These loss functions can be evaluated without any ground-truth.
Note that we omit the timestamp or frame index in some terms (e.g., $L$ instead of $L(t)$) for simplicity.

\heading{Smoothness losses.}
We measure the $C^0$ and $C^1$ smoothness of the virtual camera poses by
\begin{align}
    \coststable\!&=\!|| R_v(t) - R_v(t-\Delta t) ||^2, \\
    \costsmooth\!&=\!|| R_v(t) R_v^{-1}(t\!-\!\Delta t) - R_v(t\!-\!\Delta t) R_v^{-1}(t\!-\!2\Delta t) ||^2.
\end{align}
These two losses encourage the virtual camera to be stable and vary smoothly.

\heading{Protrusion loss.}
To avoid undefined regions and excessive cropping on the stabilized video, we measure how the warped frame protrudes the real frame boundary~\cite{shi2019steadiface}:
\begin{equation}
    \costprotrusion\!=\!\sum_{i=0}^{N} w_{p,i}||\mathtt{Min}(\mathtt{protrude}(P_v(t), P_r(t\!+\!i\Delta t))\!/\!\alpha, 1)||^2,
\end{equation}
where $N$ is the number of look-ahead frames, $w_{p,i}$ is the normalized Gaussian weights (with a standard deviation $\sigma$) centered at the current frame, and $\alpha$ is a reference protrusion value that we can tolerate.
To evaluate $\mathtt{protrude()}$, we project the virtual frame corners cropped with a ratio $\beta$ on each side to the real camera space using~\eqnref{projection} and measure the max normalized signed distance between the four warped corners to the frame boundary cropped with a ratio $\gamma$. We use $\beta$ and $\gamma$ to control the $L_p$'s sensitivity to the camera motion. The protrusion value is clamped to 1 to disregard very large motions and make training stable. 
We set $\sigma=2.5$, $N = 10$, $\alpha = 0.2$, $\beta = 0.08$ and $\gamma = 0.04$ in our experiments.

\heading{Distortion loss.}
We measure the warping distortion by:
\begin{align}
    \costdistortion &= 
    %(\mathtt{logistic}(\Omega(R_v, R_r))^2 \nonumber \\
     \Omega(R_v, R_r)  / (1 + e^{(-\beta_1 (\Omega(R_v, R_r) - \beta_0)}),
\end{align} 
where $\Omega(R_v, R_r)$ is the spherical angle between the current virtual and real camera poses, $\beta_0$ is a threshold and $\beta_1$ is a parameter to control the slope of the logistic function.
This loss is only effective when the angle deviation is larger than a threshold.
We empirically set $\beta_0 = 6^{\circ}$ and $\beta_1 = 100$ in our experiments.

\heading{Optical flow loss.}
We adopt an optical flow loss similar to~\cite{yu2019robust} to minimize the pixel motion between adjacent frames.
As shown in~\figref{flow_loss}, let $x_{r,n}$ and $y_{r,n+1}$ be the correspondences between frame $n$ and $n+1$ in the real camera space.
We define the transform from the real camera space to the virtual camera space as $T$, and obtain $x_{v,n} = T_n(x_{r,n})$ and $y_{v,n+1} = T_{n+1}(y_{r,n+1})$ in the virtual camera space.
By incorporating the forward flow $\flow{n}{n+1}$ and backward flow $\flow{n+1}{n}$, the warped pixels can be represented as:
\begin{align}
    x_{v,n} &= T_n(x_{r,n}) = T_n(y_{r,n+1} + \flow{n+1}{n}), \\
    y_{v,n+1} &= T_{n+1}(y_{r,n+1}) = T_{n+1}(x_{r,n} + \flow{n}{n+1}).
\end{align}
Our goal is to minimize $|| x_{v,n} - y_{v,n+1} ||^2$ so they stay close in the stabilized video. This can be measured by:
\begin{align}
    \label{Eq:flow_loss}
    \costflow 
    = |X_n|^{-1} \sum_{X_n} || x_{v,n} - T_{n+1}(x_{r,n} + \flow{n}{n+1}) ||^2 + \nonumber \\
      |X_{n+1}|^{-1} \sum_{X_{n+1}} || y_{v,n+1} - T_n(y_{r,n+1} + \flow{n+1}{n}) ||^2,
\end{align}
where $X_{n}$ is the set of all pixel positions in frame $n$ except those fall into undefined regions after warping.

\heading{Overall loss.}
Our final loss at a timestamp $t$ is the weighted summation of the above loss terms:
\begin{equation} \label{Eq:LossFunction}
    \cost = \weightstable\coststable + \weightsmooth\costsmooth + \weightprotrusion\costprotrusion + \weightdistortion\costdistortion + \weightflow\costflow,
\end{equation}
where $\weightstable, \weightsmooth, \weightprotrusion, \weightdistortion$ and $\weightflow$ are set to $2, 40, 2, 1$ and $1$ respectively in our experiments.

At each training iteration, we forward a sub-sequence with 100 frames to evaluate the losses and accumulate gradients before updating the model parameters.

\subsection{Multi-Stage Training} \label{Sec:MutliStageTraining} 
For the virtual camera poses, there is a trade-off between following the real camera motion and staying stable.
Although we have defined loss terms in~\eqnref{LossFunction} to constrain the solution space, it is difficult for the network to learn this non-linearity - the training cannot converge when we optimize all the loss terms simultaneously.

We adopt a multi-stage training to address this issue.
In the first stage, we only minimize $\coststable$, $\costsmooth$, and $\costdistortion$ to ensure that our model can generate a meaningful camera pose.
In the second stage, $\costprotrusion$ is added to reduce the undefined regions in the output.
In the last stage, $\costflow$ is included to enhance the overall quality.
We train each stage for 200, 100, and 500 iterations.
To improve the model generalization, we adopt a data augmentation by randomly changing the virtual camera poses (within $\pm6$ degrees) to model possible real-virtual pose deviations in the test sequences.
\vspace{-0.1in}

\begin{table}
    \centering
    \footnotesize
    \begin{tabular}{l|cccc}
        \toprule
        Method & 
        Stability & 
        Distortion & 
        Correlation & FOV \\
        \midrule
        YouTube stabilizer                          & 0.834 & \first{0.978} & 0.969 & \second{0.977} \\
        Grundmann et al.~\cite{Grundmann-CVPR-2011} & 0.818 & 0.896 & 0.948 & 0.635 \\
        Wang et al.~\cite{wang2018deep}             & 0.836 & 0.850 & 0.877 & 0.753 \\
        PWStableNet~\cite{zhao2020pwstablenet}      & 0.830 & \second{0.965} & \second{0.973} & 0.934 \\
        Yu et al.~\cite{yu2020learning}             & {0.842} & 0.854 & 0.941 & 0.793 \\
        Choi et al.~\cite{choi2020deep}             & 0.781 & 0.875 & 0.916 & \first{1.0}$^{\ast}$ \\
        Ours (sensor only)                          & \second{0.846} & 0.888 & \first{0.976} & 0.827 \\
        Ours (sensor + flow)                   & \first{0.853} & 0.937 & \first{0.976} & 0.906 \\
        \bottomrule
    \end{tabular}
    \tabmargin
    \caption{\textbf{Quantitative results.} The best one is marked in \first{red} and the second best one is marked in \second{blue}. $^{\ast}$The FOV ratio of Choi et al.~\cite{choi2020deep} is always 1 as they generate full-frame results. For other approaches, the FOV ratio is computed from the scale components of the fitted homography between the input and stabilized frames.}
    \label{tab:quantitative}
    \tabmargin
\end{table}

\section{Experimental Results} \label{Sec:results}
In this section, we show that our deep-FVS achieves state-of-the-art results in quantitative analysis (\secref{comparison}) and a user study (\secref{Userstudy}).
We then validate the effectiveness of the key components in the proposed framework by ablation study (\secref{AblationStudy}). 
\begin{figure}
    \centering
    \scriptsize
    \renewcommand{\tabcolsep}{0pt} % adjust horizontal space
	\renewcommand{\arraystretch}{0.7} % adjust vertical space
	\newcommand{\imageheight}{0.15\linewidth}
	\newcommand{\imagewidth}{0.15\linewidth}
	\newcommand{\listcase}[2]{
        \raisebox{2.4em}{\rotatebox[origin=c]{90}{#1}} &
        \includegraphics[height=\imageheight]{figures/metrics/#2/Axis_crop.png} &  
        \includegraphics[height=\imageheight]{figures/metrics/#2/General_crop.png} &  
        \includegraphics[height=\imageheight]{figures/metrics/#2/Rotation_crop.png} &  
        \includegraphics[height=\imageheight]{figures/metrics/#2/Parallax_crop.png} &  
        \includegraphics[height=\imageheight]{figures/metrics/#2/Driving_crop.png} &  
        \includegraphics[height=\imageheight]{figures/metrics/#2/People_crop.png} &  
        \includegraphics[height=\imageheight]{figures/metrics/#2/Running_crop.png}
	}
    \begin{tabular}{lccccccc}
        & & 
        \includegraphics[width=\imagewidth]{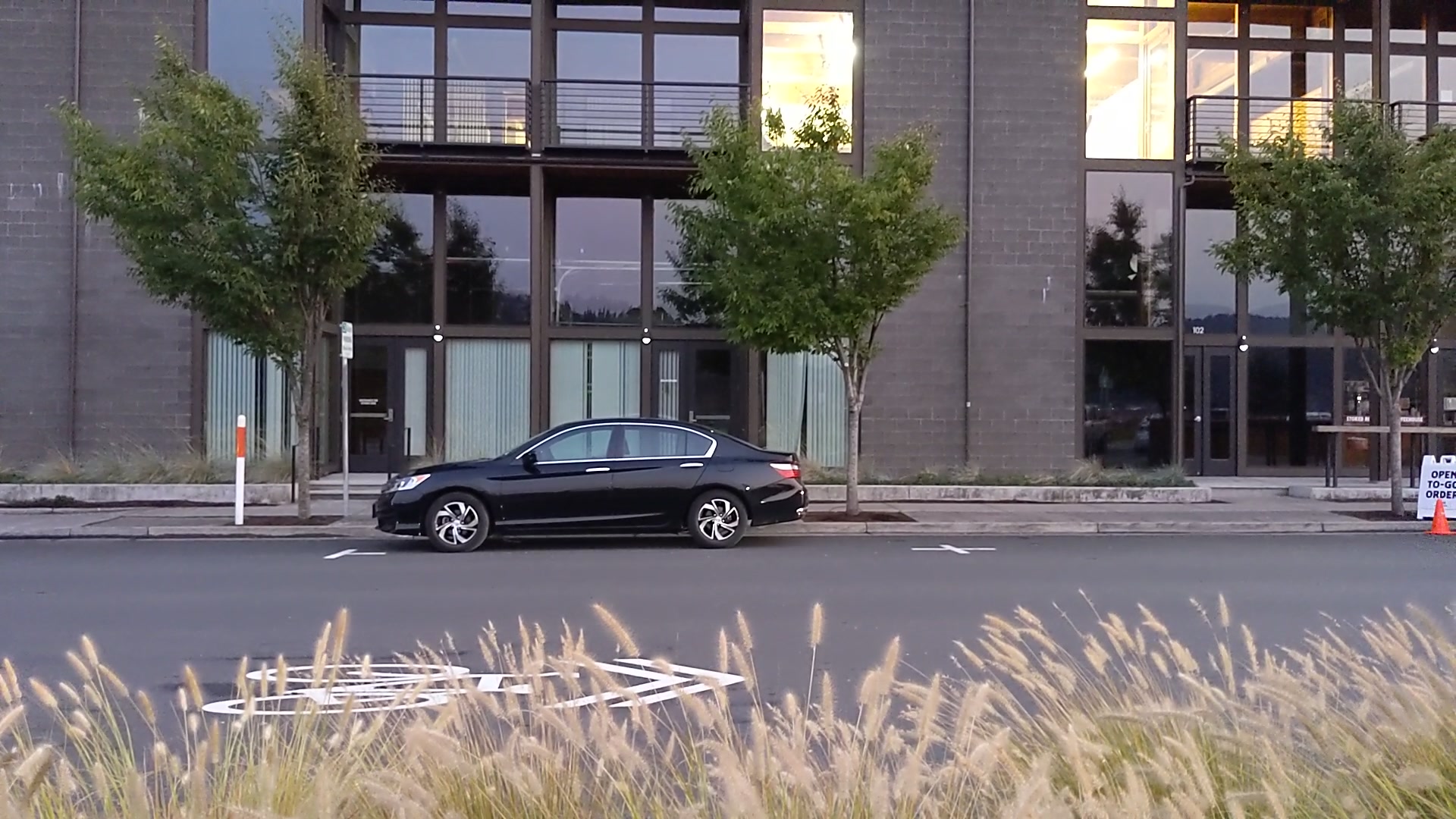} &
        \includegraphics[width=\imagewidth]{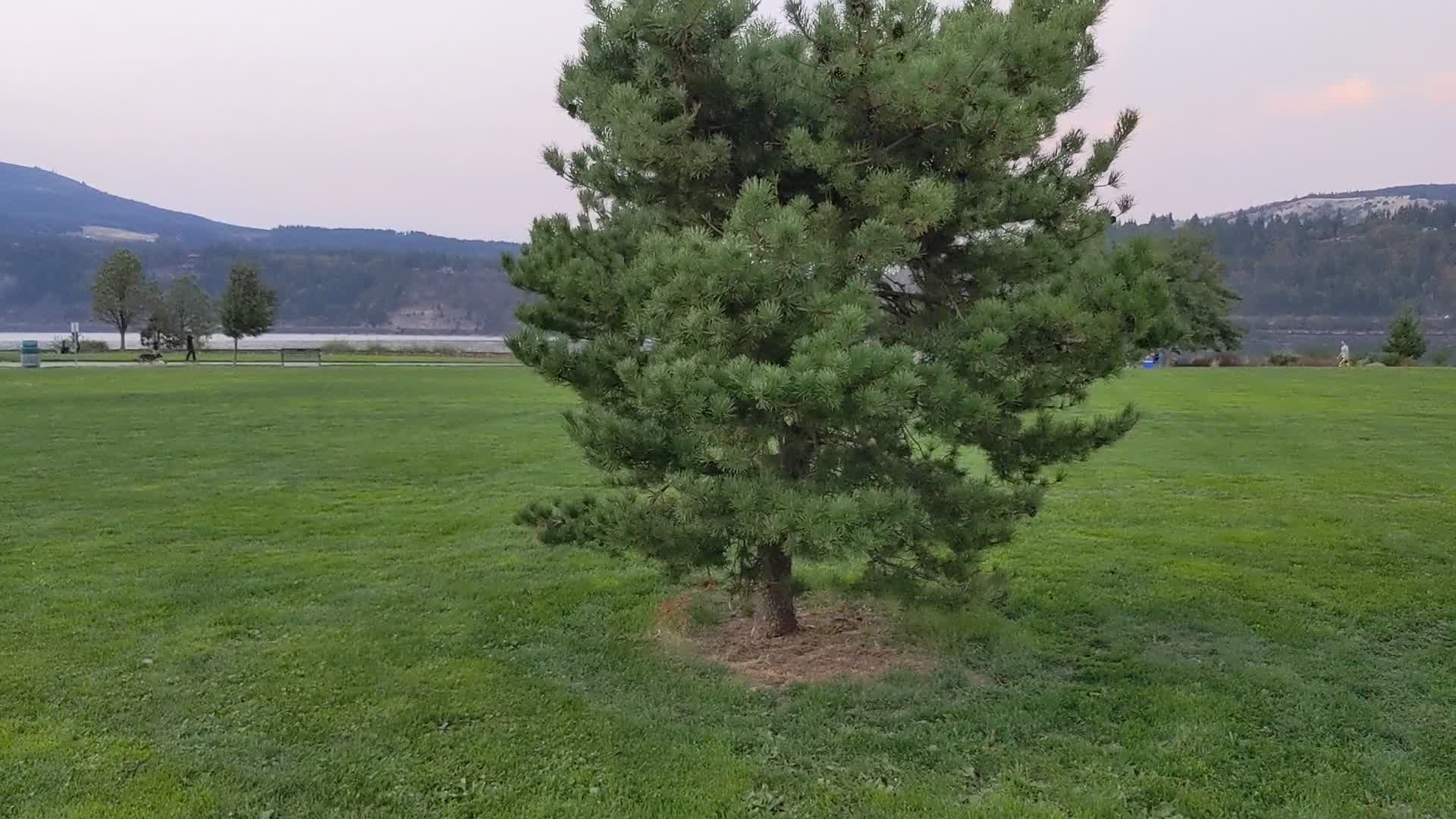} &
        \includegraphics[width=\imagewidth]{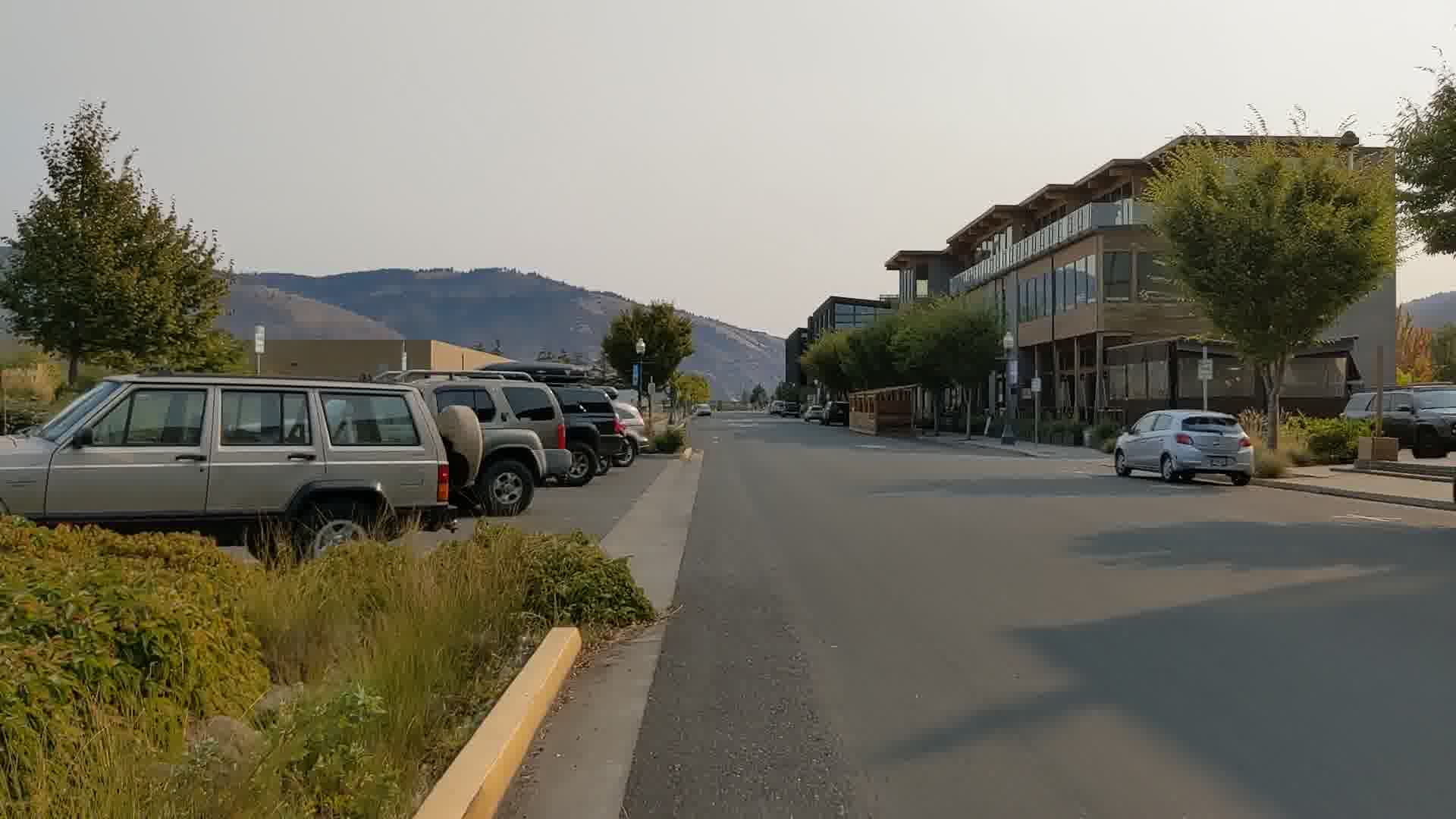} &
        \includegraphics[width=\imagewidth]{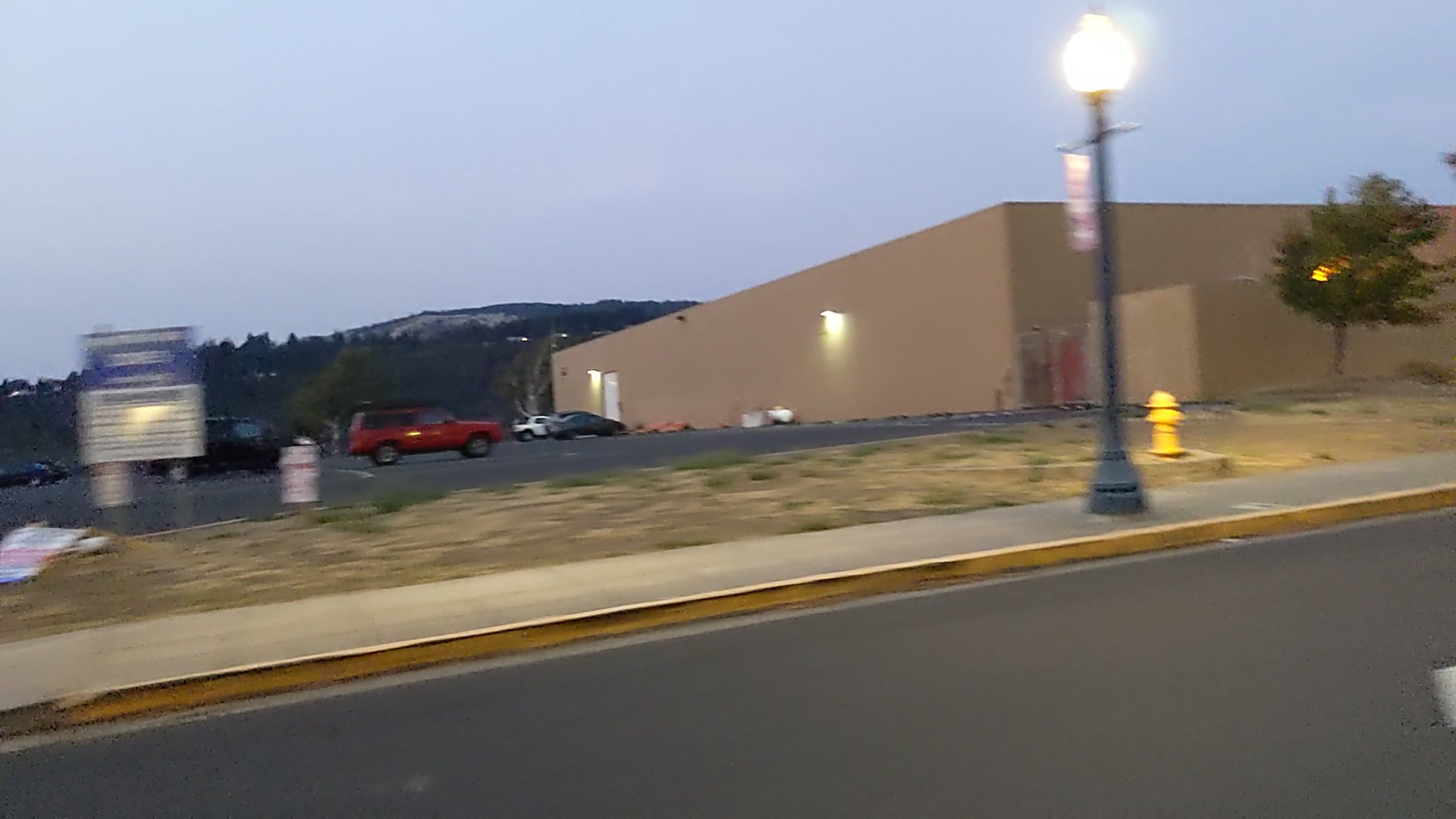} &
        \includegraphics[width=\imagewidth]{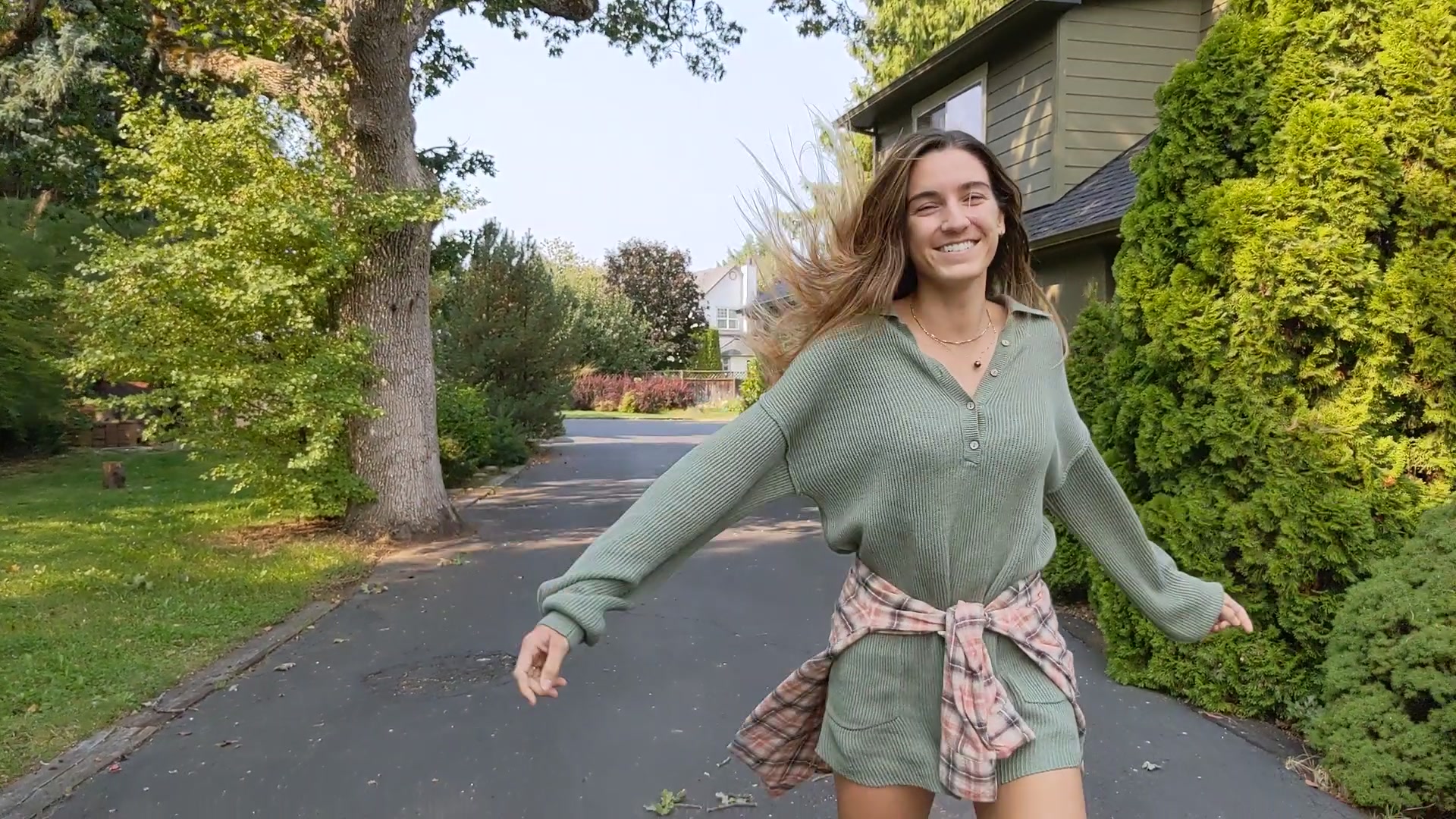} &
        \includegraphics[width=\imagewidth]{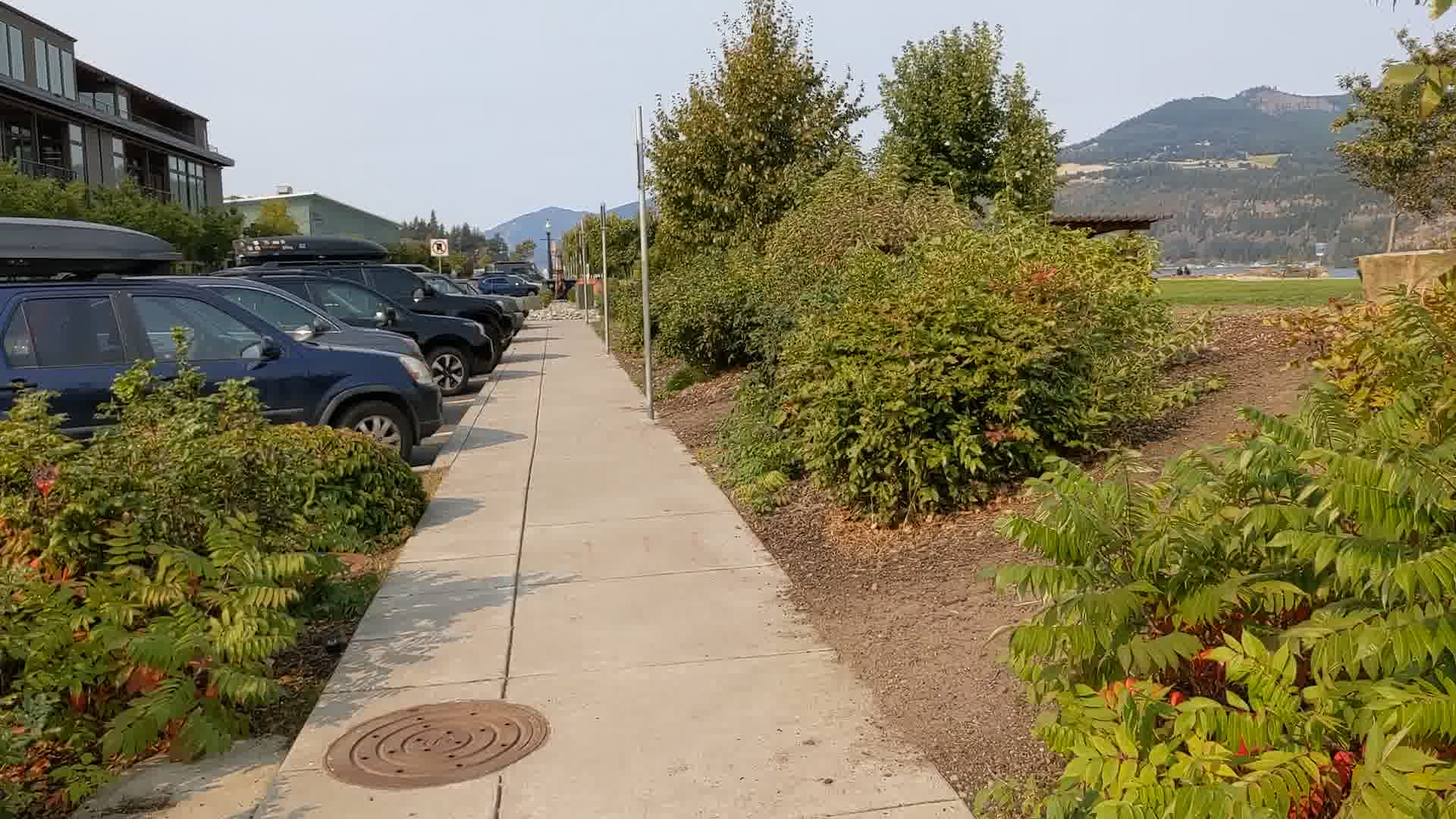} 
        \\
        & & \textsc{General} & \textsc{Rotation} & \textsc{Parallax} & \textsc{Driving} & \textsc{People} & \textsc{Running} 
        \\
        \listcase{Stability}{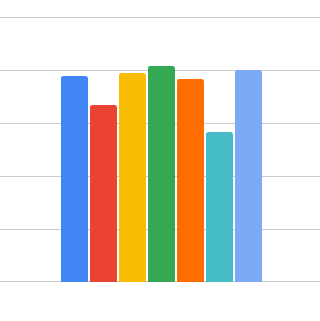}  \\
        \listcase{Distortion}{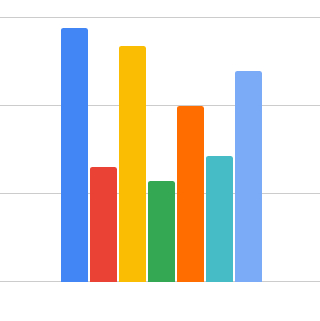}  \\
        \listcase{Correlation}{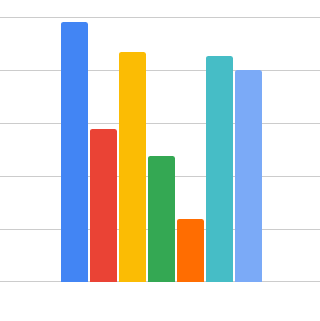}  \\
        \listcase{FOV ratio}{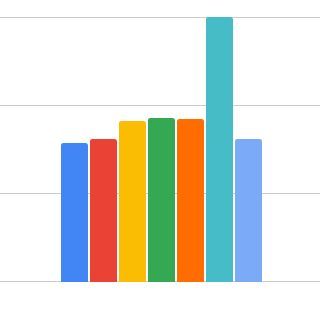}  \\
        & & \multicolumn{6}{c}{\includegraphics[width=0.7\linewidth]{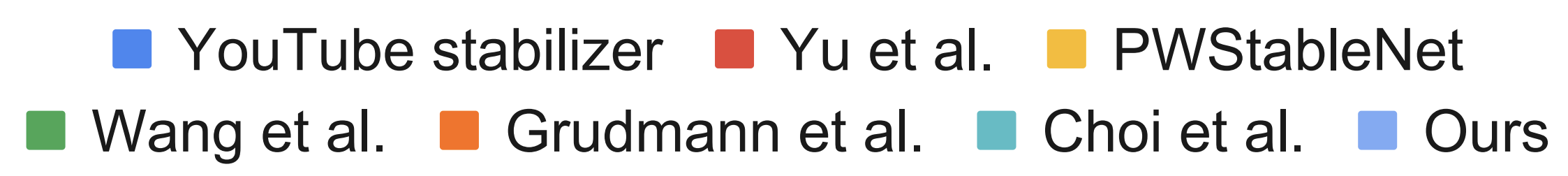}}
    \end{tabular}
    \figmargin
    \caption{\textbf{Per-category quantitative evaluation.}
    We compare the stability, FOV ratio, distortion, and correlation with state-of-the-art methods~\cite{choi2020deep,Grundmann-CVPR-2011,wang2018deep,yu2020learning,zhao2020pwstablenet} on each category. 
    }
    \label{fig:category_metric}
    \figmargin
\end{figure}
We strongly encourage readers to watch the source and stabilized videos (by our and existing methods) in the supplemental materials.

\begin{figure*}
    \centering
    \footnotesize
    \renewcommand{\tabcolsep}{1pt} % adjust horizontal space
	\renewcommand{\arraystretch}{0.8} % adjust vertical space
	\newcommand{\imagewidth}{0.24\linewidth}
    \begin{tabular}{cccc}
        \includegraphics[width=\imagewidth]{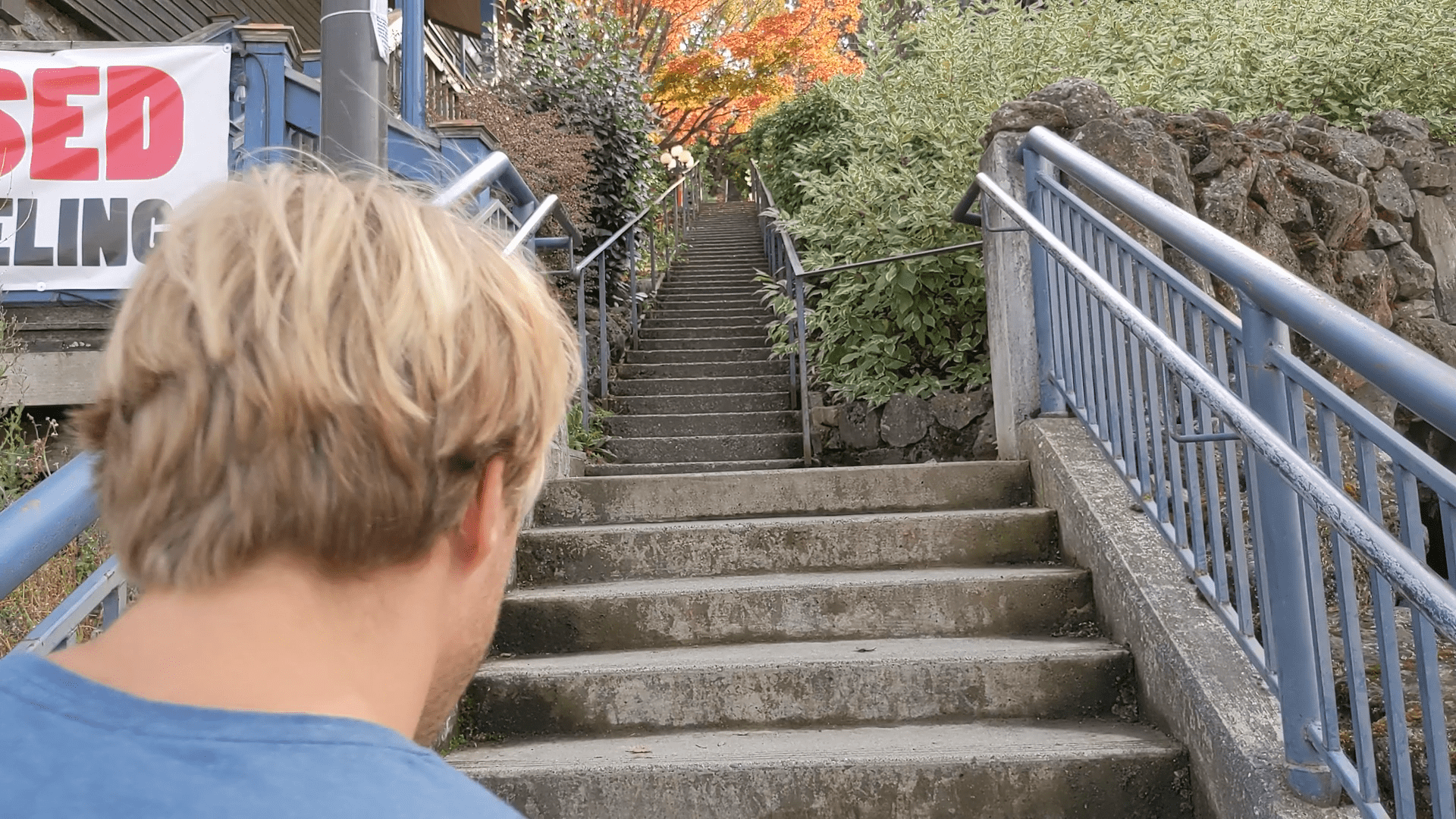} &
        \includegraphics[width=\imagewidth]{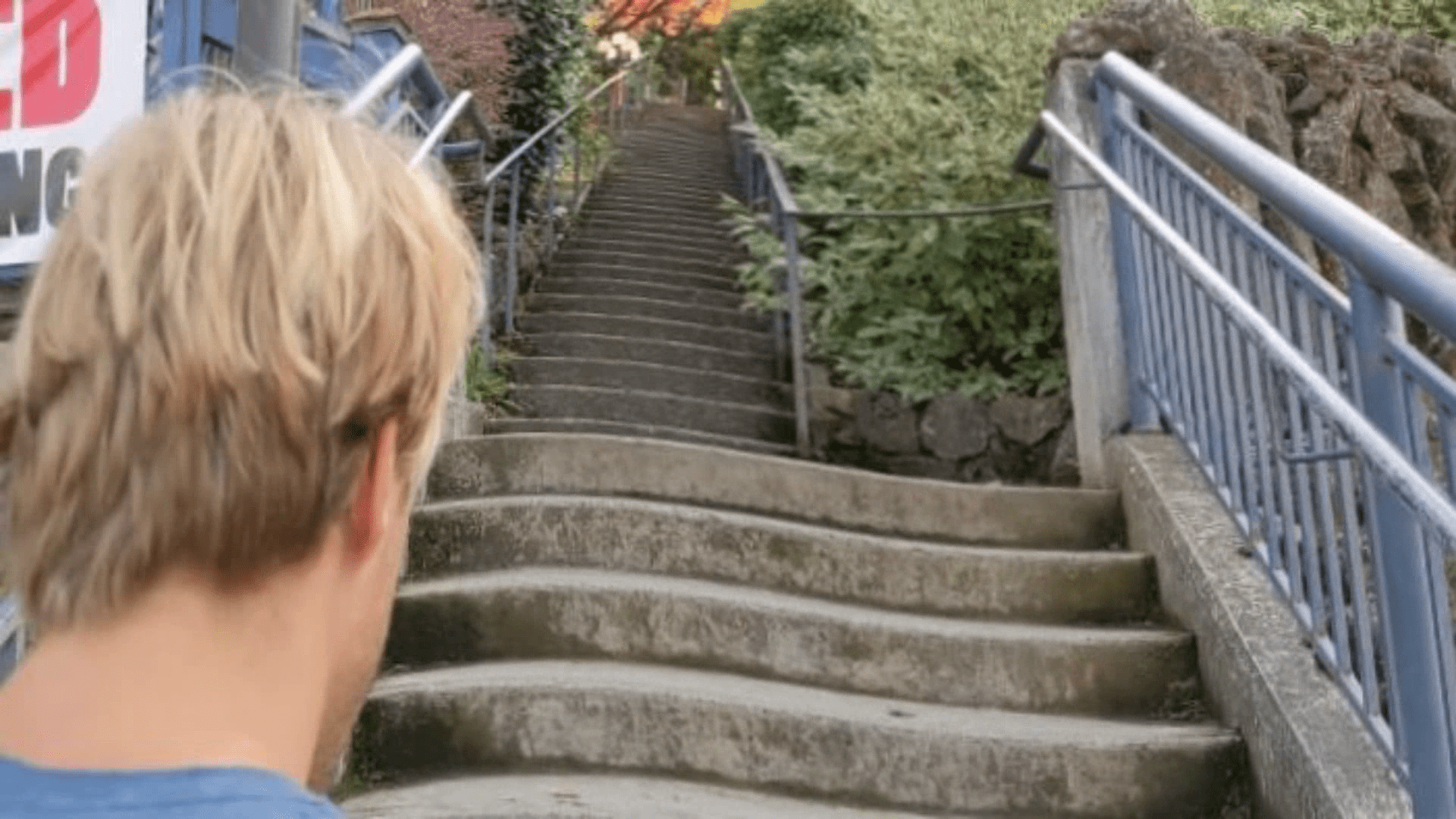} &
        \includegraphics[width=\imagewidth]{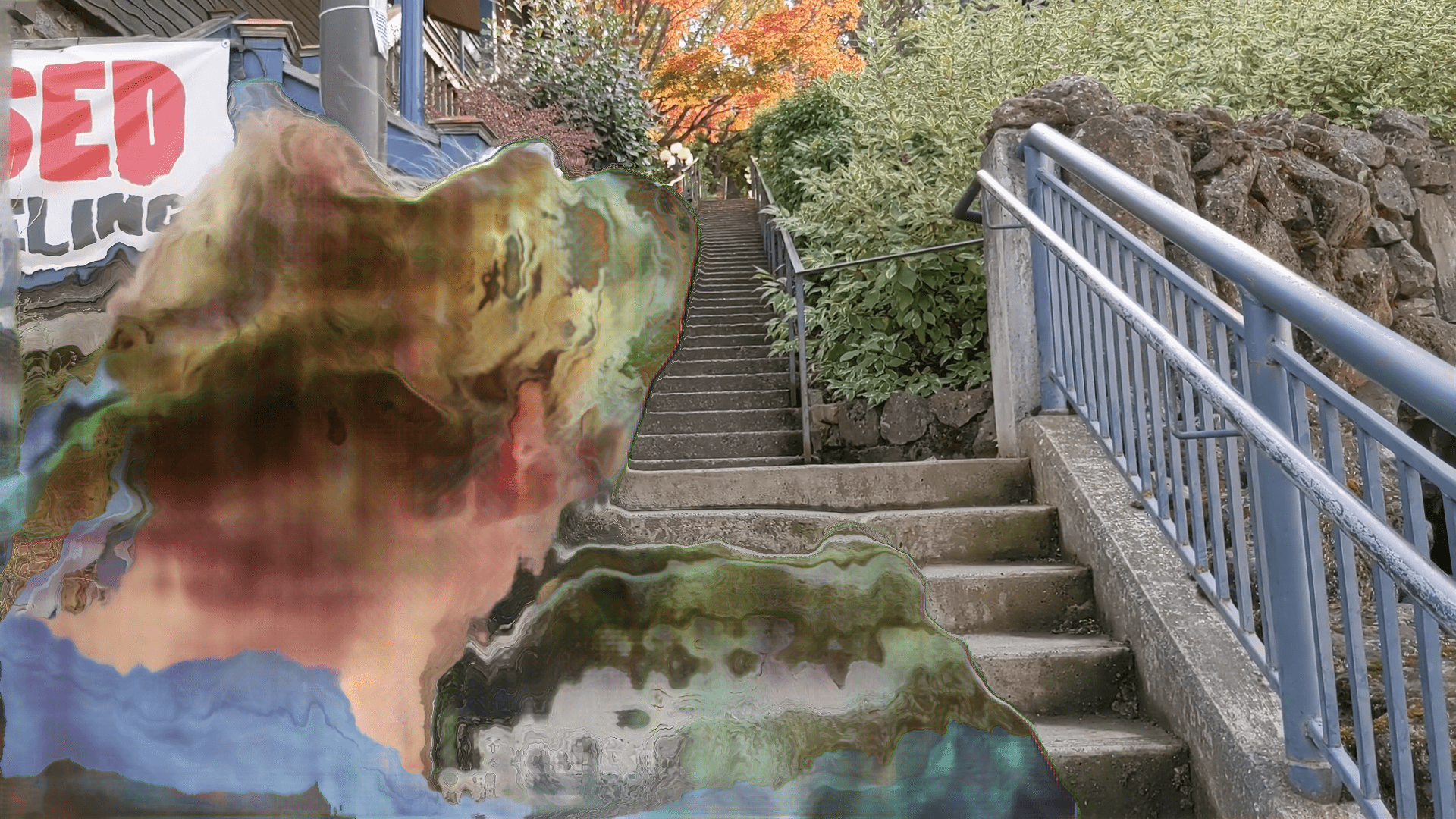} &
        \includegraphics[width=\imagewidth]{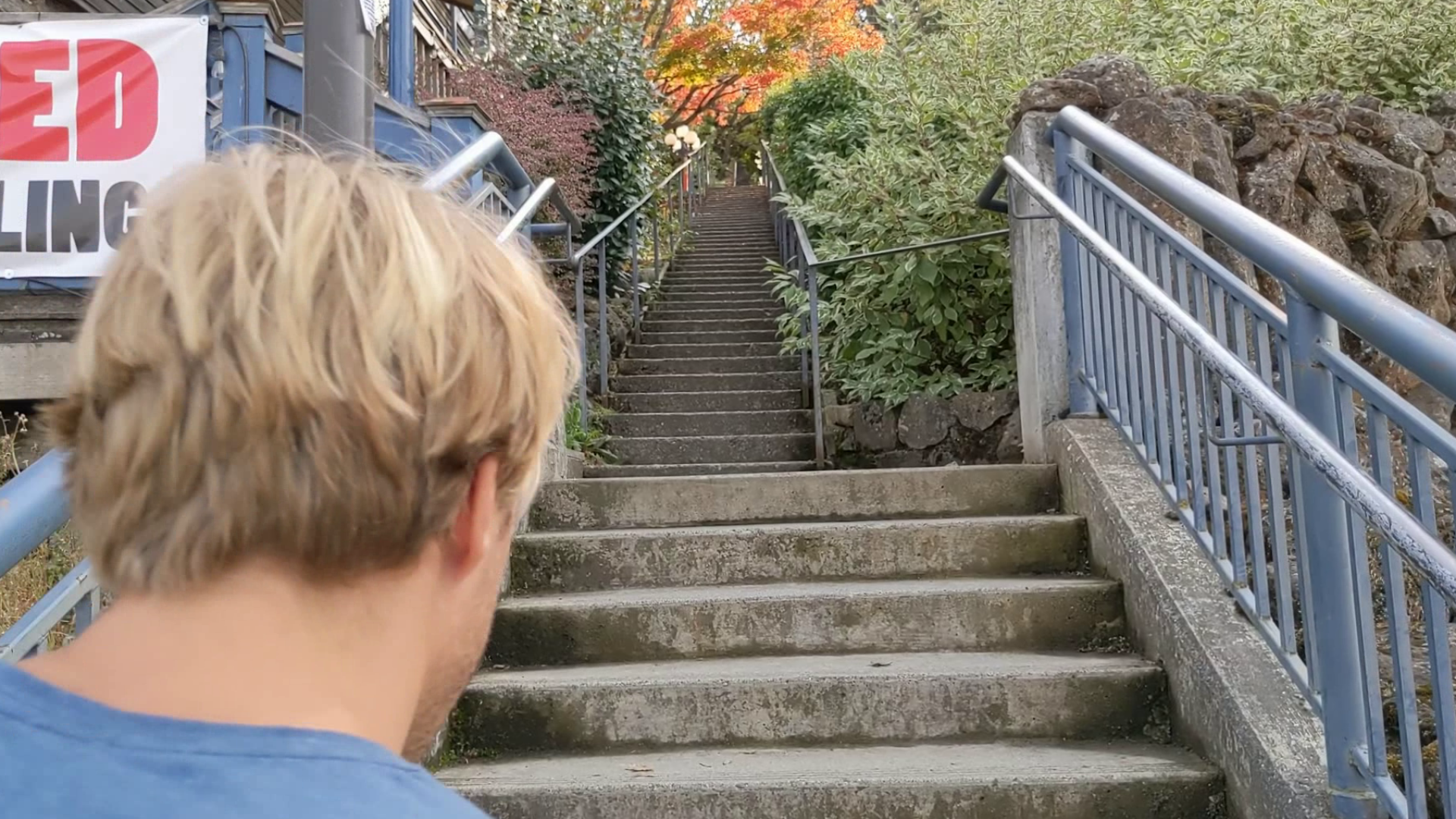} \\
        Input frame &
        Yu et al.~\cite{yu2020learning} &
        Choi et al.~\cite{choi2020deep} &
        Ours \\
        \includegraphics[width=\imagewidth]{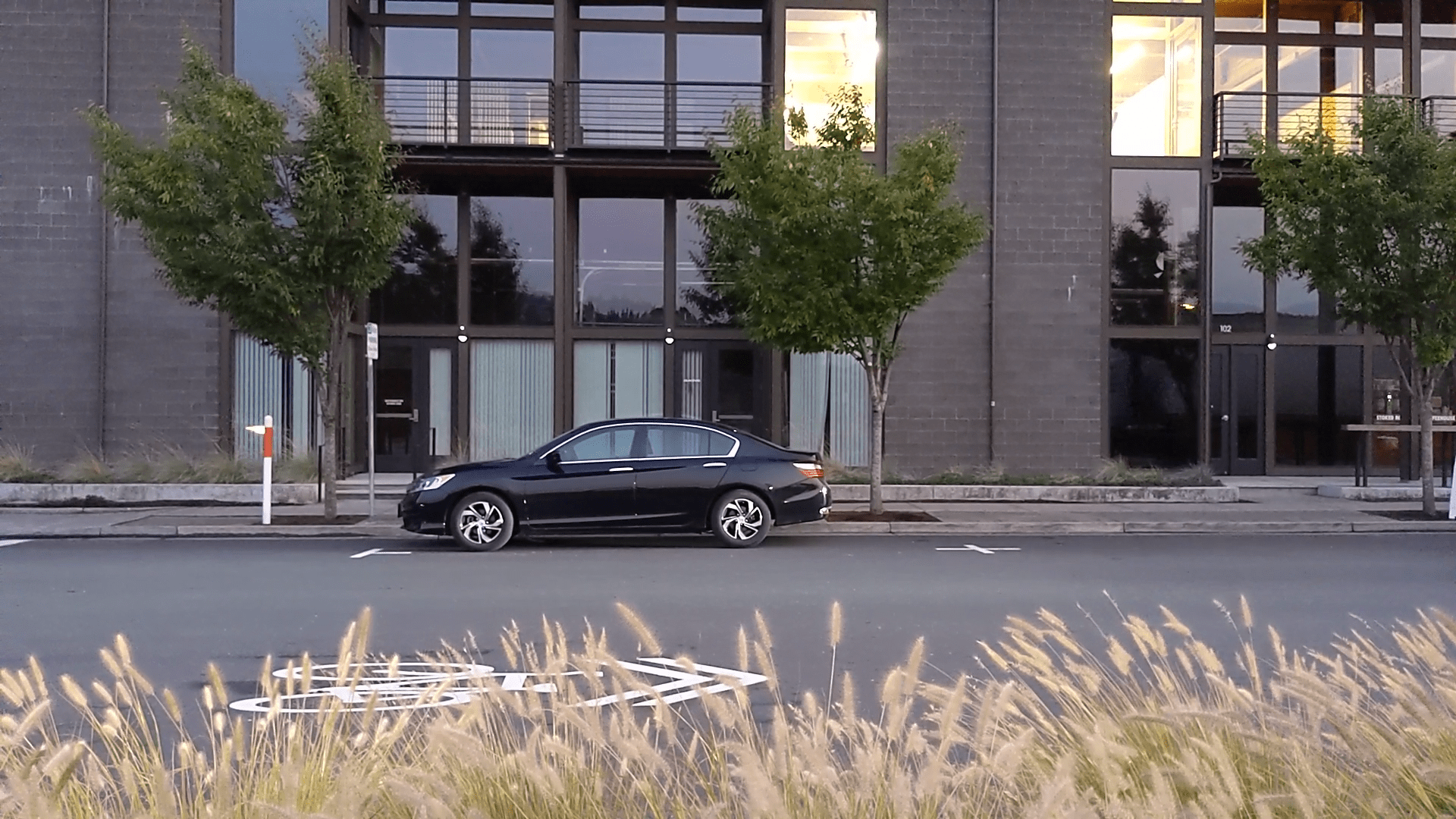} &
        \includegraphics[width=\imagewidth]{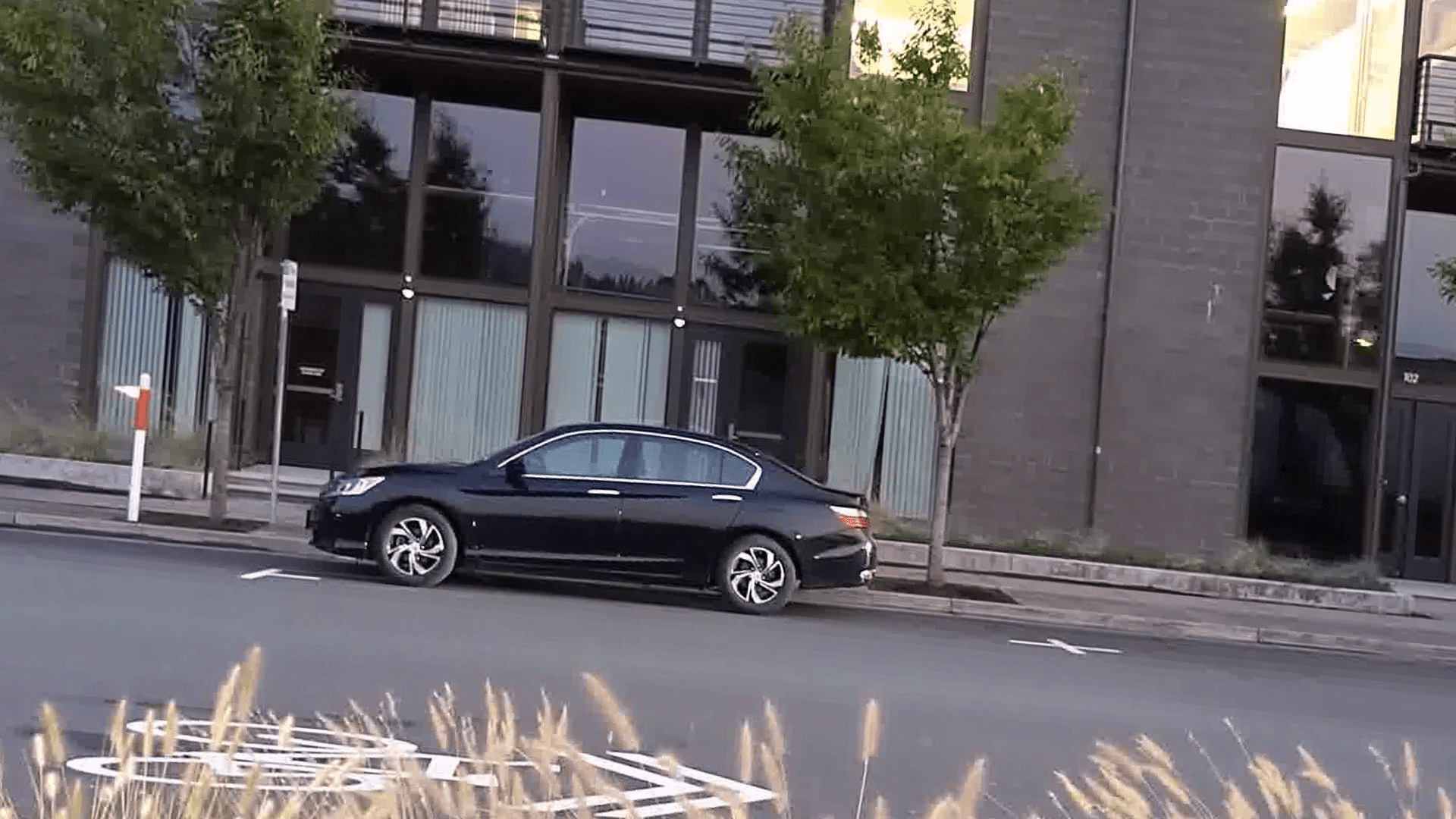} &
        \includegraphics[width=\imagewidth]{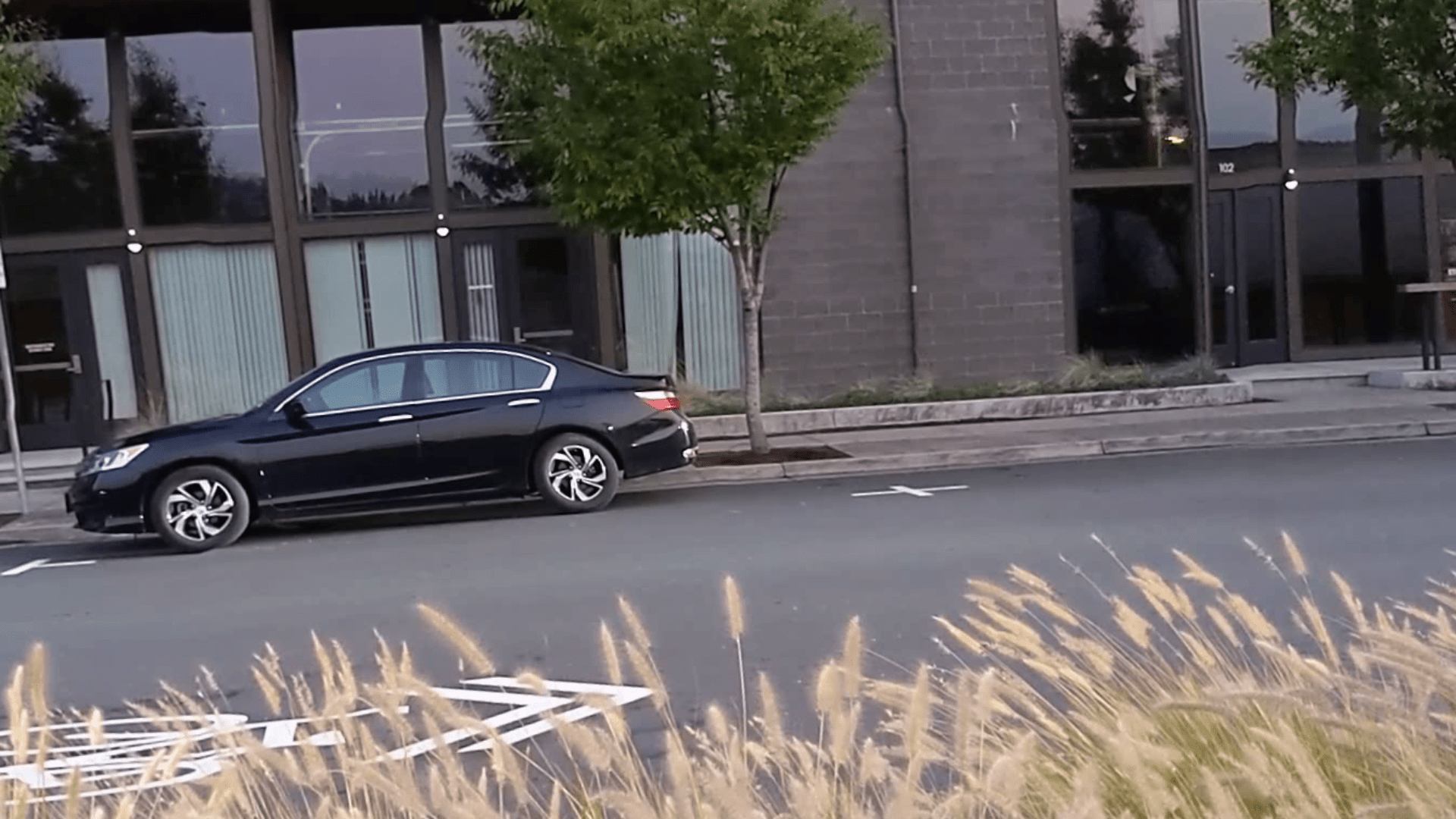} &
        \includegraphics[width=\imagewidth]{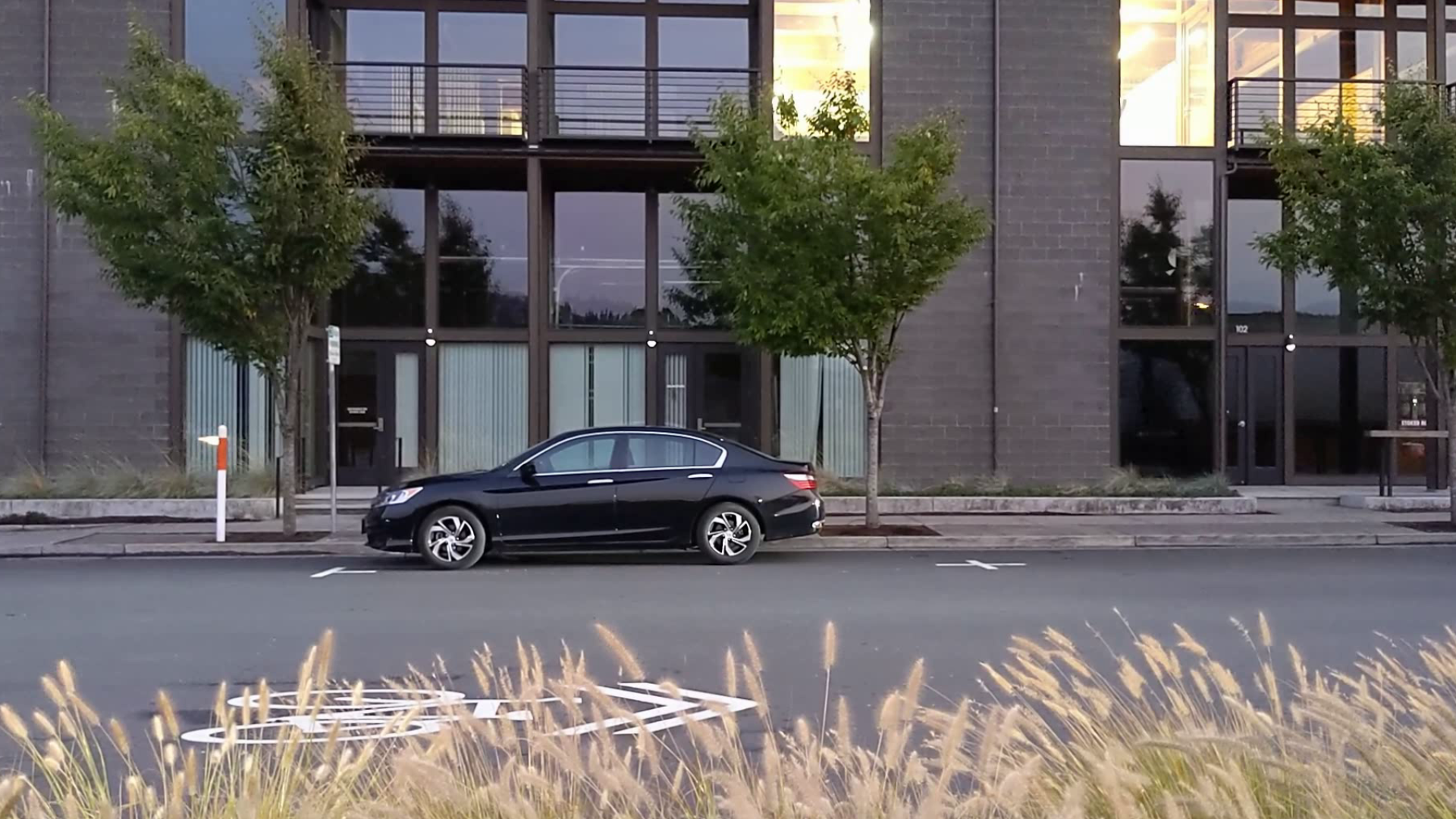} \\
        Input frame &
        Grundmann et al.~\cite{Grundmann-CVPR-2011} &
        Wang et al.~\cite{wang2018deep} &
        Ours \\
    \end{tabular}
    \figmargin
    \caption{\textbf{Visual comparisons.}
    Non-rigid distortion, local artifacts and temporal wobbling are observed in Yu et al.~\cite{yu2020learning} and Choi et al.~\cite{choi2020deep}, and large rotation deviation observed in Grundmann et al.~\cite{Grundmann-CVPR-2011} and Wang et al.~\cite{wang2018deep} (which also exhibits local distortions). Our method is free of such issues.
    Please refer to our supplemental videos on the full results of all the methods.
    }
    \label{fig:visual_comparison}
    \figmargin
    % \vspace{1mm}
\end{figure*}

\begin{figure*}
    \centering
    \footnotesize
    \renewcommand{\tabcolsep}{1pt} % adjust horizontal space
	\renewcommand{\arraystretch}{0.8} % adjust vertical space
	\newcommand{\imagewidth}{0.24\linewidth}
	\newcommand{\listcase}[1]{
	    \includegraphics[width=\imagewidth]{figures/comparisons/stay/#1/origin_zoom.png} & 
        \includegraphics[width=\imagewidth]{figures/comparisons/stay/#1/MAT_zoom.png} & 
        \includegraphics[width=\imagewidth]{figures/comparisons/stay/#1/Tsing_zoom.png} & 
        \includegraphics[width=\imagewidth]{figures/comparisons/stay/#1/PWS_zoom.png} \\
        Input frame &
        Grundmann et al.~\cite{Grundmann-CVPR-2011} &
        Wang et al.~\cite{wang2018deep} &
        PWStableNet~\cite{zhao2020pwstablenet} \\
        \includegraphics[width=\imagewidth]{figures/comparisons/stay/#1/ytb_zoom.png} &
        \includegraphics[width=\imagewidth]{figures/comparisons/stay/#1/Jiyang_zoom.png} & 
        \includegraphics[width=\imagewidth]{figures/comparisons/stay/#1/DIFRINT_zoom.png} & 
        \includegraphics[width=\imagewidth]{figures/comparisons/stay/#1/our_zoom.png} \\
        YouTube stabilizer &
        Yu et al.~\cite{yu2020learning} &
        Choi et al.~\cite{choi2020deep} &
        Ours
	}
    \begin{tabular}{cccc}
        \listcase{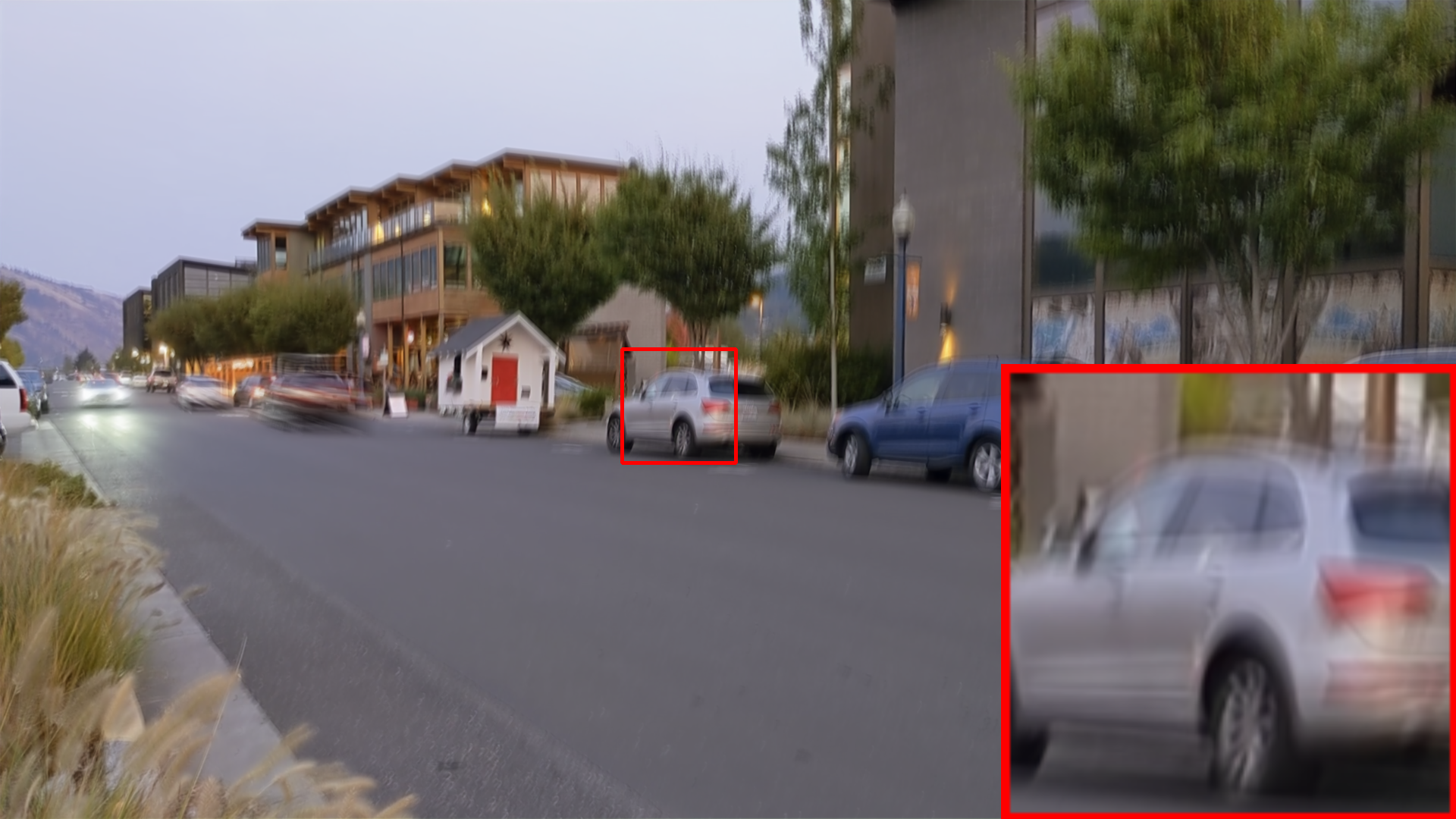}
    \end{tabular}
    \figmargin
    \caption{\textbf{Stability comparisons.} 
    We take a video with almost no camera motion except handshakes and average 11 adjacent frames.
    Our average frame is sharper than other methods, indicating that our result is more stable.
    Please refer to our supplemental videos on the full results of all the methods.
    }
    \label{fig:stability}
    \figmargin
\end{figure*}

\subsection{Comparisons with State-of-the-Arts} \label{Sec:comparison}

\heading{Experimental settings.}
We compare our deep-FVS with conventional methods~\cite{Grundmann-CVPR-2011}\footnote{We use a third-party implementation from \nolinkurl{https://github.com/ishit/L1Stabilizer}.} and YouTube stabilizer (based on~\cite{Grundmann2012calibration}), and 4 recent learning-based methods~\cite{choi2020deep,wang2018deep,yu2020learning,zhao2020pwstablenet}\footnote{The source code of~\cite{choi2020deep,wang2018deep,zhao2020pwstablenet} are publicly available. We obtain the source code of~\cite{yu2020learning} from the authors.}.
We collect 50 videos with sensor logs using Google Pixel 4 ($1920 \times 1080$ resolution with variable FPS).
The video dataset covers a wide range of variations, such as scenes, illuminations, and motions.
The sensor data are accurately calibrated and timestamp-aligned with frames (Google Pixel 4 is ARCore certified~\cite{arcore_certificate}).
We split our dataset into 16 videos for training and 34 videos for testing, where the test set classified into 6 categories: \textsc{general}, \textsc{rotation}, \textsc{parallax}, \textsc{driving}, \textsc{people}, and \textsc{running}.
\figref{category_metric} shows a few sample frames from each category.

\heading{Quantitative comparisons.}
We use three commonly used metrics~\cite{choi2020deep,Liu-TOG-2013,wang2018deep,yu2020learning,zhao2020pwstablenet}: \emph{Stability}, \emph{Distortion}, and \emph{FOV ratio}, and define a \emph{Correlation} score, to evaluate the tested methods (please refer to the supplemental materials on their definitions).
Note that the distortion measures the global geometry deviation from the input frames, while the correlation evaluates the local deformation.

The results of all test videos are summarized in~\tabref{quantitative}, and \figref{category_metric} plots the average scores for the 6 categories.
Overall, our method achieves the best stability and correlation scores.
The YouTube stabilizer and Choi et al.~\cite{choi2020deep} generate nearly uncropped or full-frame results, while our FOV ratio is comparable to PWStableNet~\cite{zhao2020pwstablenet}.
We found the quantitative gaps in existing metrics do not fully reflect large visual differences. For example, YouTube stabilizer keeps frames nearly uncropped but outputs relatively unstable videos. 
PWStableNet~\cite{zhao2020pwstablenet} produces lots of residual global motions and temporal wobbling, which are not captured by the distortion score.
Our method generally obtains better stability and correlation scores on challenging \textsc{rotation}, \textsc{running}, and \textsc{people} categories.

\heading{Qualitative comparisons.}
We provide visual comparisons of stabilized frames in~\figref{visual_comparison}.
Both Yu et al.~\cite{yu2020learning} and Choi et al.~\cite{choi2020deep} use optical flows to warp the frames and often generate local distortion (03:20 in demo video).
Choi et al.~\cite{choi2020deep} produce severe artifacts when the motion is large (04:55 in demo video).
Grundmann et al.~\cite{Grundmann-CVPR-2011} estimate a global transformation, and Wang et al.~\cite{wang2018deep} predict low-resolution warping grids.
The results of both methods have less local distortion but are not temporally stable as the motion is purely estimated from the video content (03:56, 06:02 in demo video).
In contrast, we fuse both the gyroscope data and optical flow for more accurate motion inference and obtain stable results without distortion or wobbling.

\figref{stability} shows the averaged frame of 11 adjacent frames from a short clip, where the input video contains only handshake motion.
Ideally, the stabilized video should look static as it was captured on a tripod.
Our result is the sharpest one, while the averaged frames from other approaches~\cite{choi2020deep, Grundmann-CVPR-2011,wang2018deep,yu2020learning, zhao2020pwstablenet} are blurry, demonstrating that our result is more stable than others (03:35 in demo video).
We highly encourage readers to watch the full video comparisons in the demo to better evaluate the stabilization quality.

\heading{User Study.}\label{Sec:Userstudy}
We conduct a user study to evaluate human preferences on the stabilized videos.
As it is easier for a user to make a judgment between two results instead of ranking multiple videos, we adopt the paired comparison~\cite{lai2016comparative, rubinstein2010comparative} to measure the subject preference.
In each test, we show two stabilized videos side-by-side and the input video as a reference.
We ask the participant the following questions: 
\begin{compactenum}
\item Which video is more stable?
\item Which video has less distortion?
\item Which video has a larger FOV?
\end{compactenum}

In total, we recruit 50 participants online, where each participant evaluates 18 pairs of videos.
The videos are shuffled randomly when presenting to each user. 
All the methods are compared the same number of times.

\tabref{userstudy} shows that our method is preferred in more than $95\%$ of comparisons regarding the stability and $93\%$ regarding the visual quality (less distortion).
Our results have larger FOVs than Grundmann et al.~\cite{Grundmann-CVPR-2011} and Wang et al.~\cite{wang2018deep} as these two approaches apply excessive cropping to avoid the irregular boundaries.
Our FOV is comparable to PWStableNet~\cite{zhao2020pwstablenet} as users cannot tell the difference ($\simeq50\%$).
Yu et al.~\cite{yu2020learning} generates results with a large FOV in most cases, but applies excessive cropping when the video motion is large (e.g., running).
Choi et al.~\cite{choi2020deep} generates full-frame results but at the cost of visible distortion.
YouTube stabilizer applies lightweight changes to preserve most of the input content, but the results are less stable.
The user study demonstrates that our method generates more stable results with fewer visual artifacts and distortions, while the amount of cropping is similar to other approaches.

\begin{table}[]
    \centering
    \resizebox{\linewidth}{!}{
    \begin{tabular}{l|c|c|c}
        \toprule
        & More stable & Less distortion & Larger FOV \\
        \midrule
vs. Grundmann et al.~\cite{Grundmann-CVPR-2011} &   98.4$\pm$2.3$\%$ & 95.9$\pm$3.5$\%$ & 82.1$\pm$6.9$\%$ \\
vs. Wang et al.~\cite{wang2018deep} &       98.4$\pm$2.3$\%$ & 95.1$\pm$3.9$\%$ & 77.2$\pm$7.5$\%$ \\
vs. PWStableNet~\cite{zhao2020pwstablenet} & 91.1$\pm$5.1$\%$ & 89.4$\pm$5.5$\%$ & 48.0$\pm$9.0$\%$ \\
vs. Yu et al.~\cite{yu2020learning} &           92.7$\pm$4.7$\%$ & 93.5$\pm$4.4$\%$ & 38.2$\pm$8.7$\%$ \\
vs. Choi et al.~\cite{choi2020deep} &        96.7$\pm$3.2$\%$ & 97.6$\pm$2.8$\%$ & 27.6$\pm$8.0$\%$ \\
vs. YouTube stabilizer &         96.7$\pm$3.2$\%$ & 88.6$\pm$5.7$\%$ & 23.6$\pm$7.6$\%$ \\
\midrule
Average &             95.7$\pm$1.5$\%$ & 93.4$\pm$1.8$\%$ & 49.5$\pm$3.6$\%$ \\
\bottomrule
    \end{tabular}
    }
    \tabmargin
    \caption{\textbf{Results of user study.} Our results are more stable with less distortion and overall a comparable field-of-view. 
    }
    \label{tab:userstudy}
    \tabmargin
\end{table}

\begin{figure}
    \centering
    \footnotesize
    \includegraphics[width=1.0\linewidth]{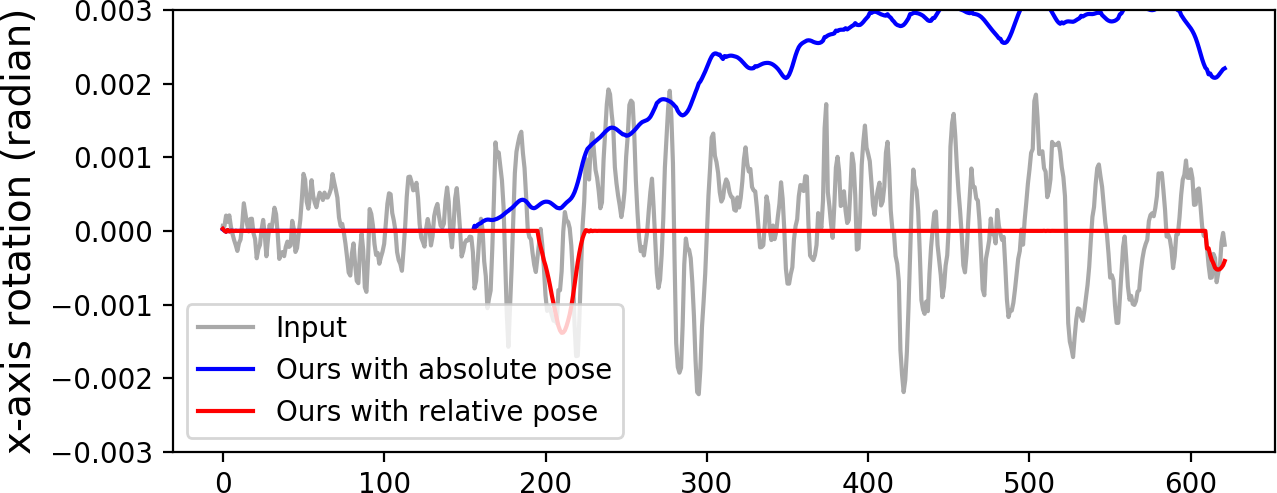}
    \\
    (a) Analysis on relative pose \\
    \includegraphics[width=1.0\linewidth]{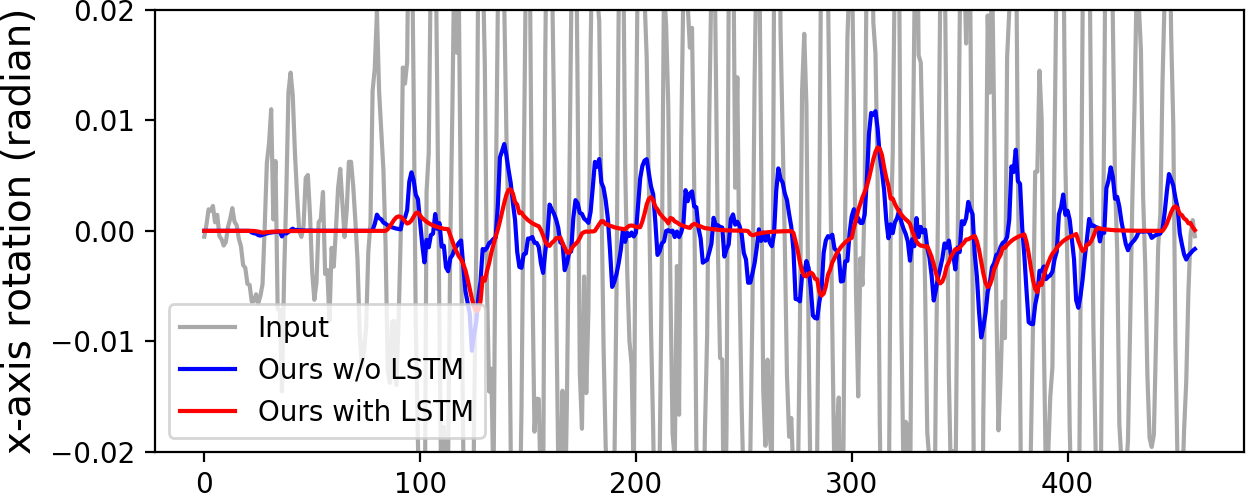}
    \\
    (b) Analysis on LSTM
    \includegraphics[width=1.0\linewidth]{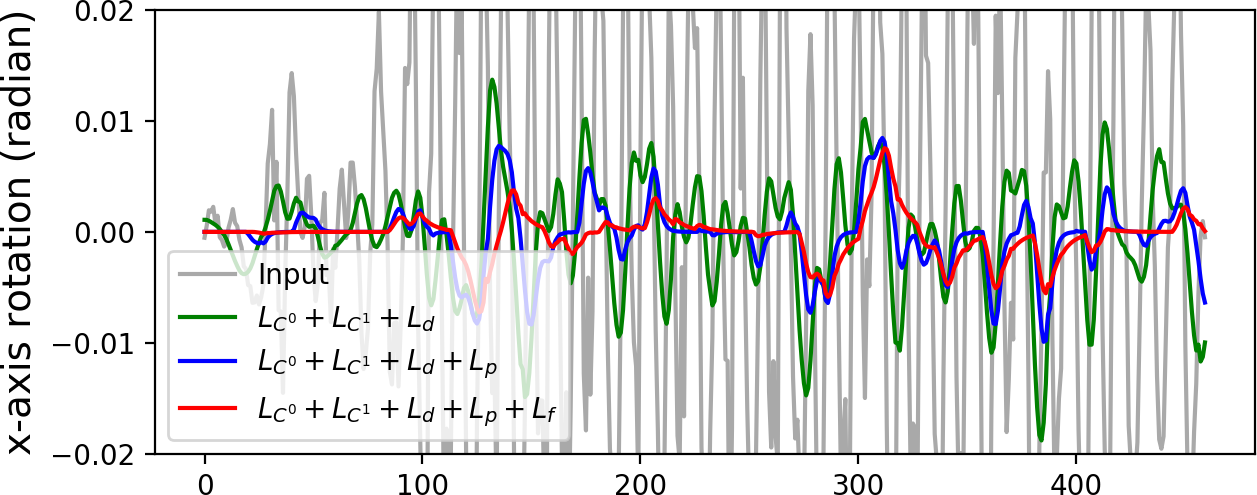}
    \\
    (c) Analysis on loss functions \\
    \figmargin
    \caption{\textbf{Ablation studies on relative poses, LSTM and losses.}
    (a) The model using relative poses can output more stable poses and follow the real camera motion well.
    (b) Without LSTM, our model cannot learn motion patterns well and often generate unstable prediction.
    (c) The protrusion loss $\costprotrusion$ reduces the undefined region, and the optical flow loss $\costflow$ further improves the smoothness.  
    }
    \label{fig:ablation}
    \figmargin
\end{figure}

\subsection{Ablation Study} \label{Sec:AblationStudy}
\heading{Importance of using relative poses.}
As the same motion patterns can be converted to similar relative poses, it is easier for the model to infer the motion pattern from rotation deviations instead of the absolute poses. 
Using the relative poses also makes the model training more numerically stable.
\figref{ablation}(a) shows that our method with relative poses can follow the real camera poses well for a \textsc{panning} case.
In contrast, the model using absolute poses deviates away from the real motion. 

\heading{Importance of LSTM.}
The LSTM unit carries the temporal information (e.g., motion state) and enables the model to output state-specific results. 
With the temporal information, the LSTM can also reduce high-frequency noise and generate more stable poses.
As shown in~\figref{ablation}(b), when replacing the LSTM with an FC layer, the output poses contain more jitter, resulting in less stable videos.

\begin{figure}
    \centering
    \footnotesize
    \includegraphics[width=1.0\linewidth]{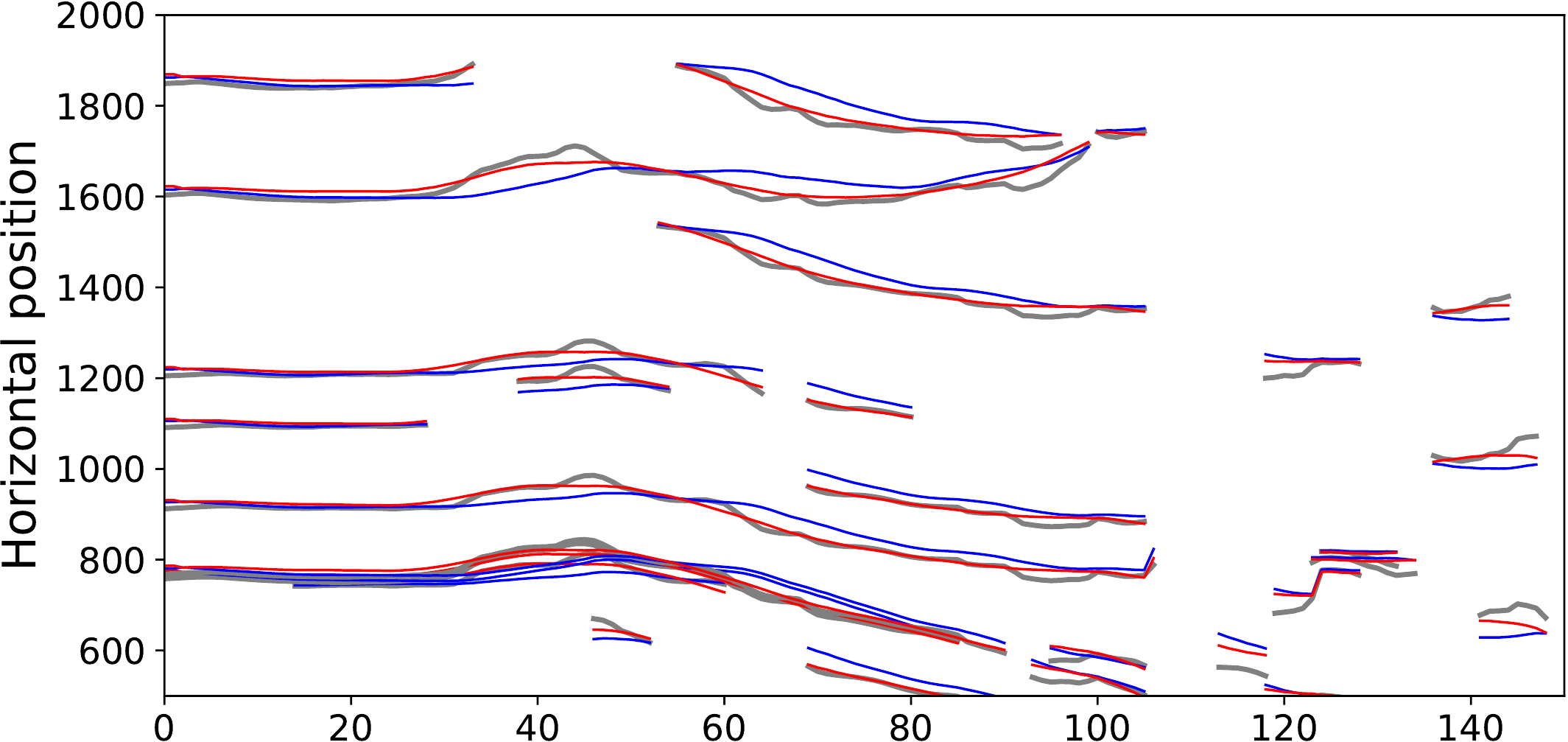}
    \figmargin
    \caption{\textbf{Feature trajectories before and after stabilization.} Grey: input video. Blue: our method with sensor only (stage 2). Red: our method with sensor and optical flow fusion (stage 3). Our final fused model produces the most stable trajectories which also better follow the real motion.
    }
    \label{fig:tracking}
    \figmargin
\end{figure}
\heading{Importance of optical flow loss.}
We compare the sensor-only stabilization results in stage 2 and the full fused results with the optical flow term $L_f$ in stage 3 (Sec.~\ref{Sec:MutliStageTraining}). 
The optical flow loss enables the model to adapt to scene content (e.g. parallax), which can improve stability and increase the FOV as shown in the last two rows of Table~\ref{tab:quantitative}.
We also visualize the feature trajectories detected from the KLT tracker~\cite{lucas1981iterative} in~\figref{tracking}.
Our method with sensor and optical flow fusion (red curves) can follow the original camera motion well and maintain the smoothness.

\heading{Importance of other losses.}
\figref{ablation}(c) shows the x-axis virtual rotation for a \textsc{running} case. We see a step-by-step stability improvement (smoother motion curves) after introducing each loss.

We also provide the comparisons of execution time in the supplementary material.

\section{Limitations and Conclusion}
In this work, we present deep Fused Video Stabilization, the first DNN-based unsupervised framework that utilizes both sensor data and images to generate high-quality distortion-free results. 
The proposed network achieves high-quality performance using joint motion representation, relative motion history, unsupervised loss functions, and multi-stage training.
We have demonstrated that our method outperforms state-of-the-art alternatives in both quantitative comparisons and user study.
We will release the source code and the video dataset to facilitate future research.

Unlike image-based methods, our framework requires additional sensor data. 
Fortunately, most modern smartphones have a well-synchronized sensor and camera system~\cite{arcore_certificate} which makes the requirement minimal. 
Our model does not support hard FOV preservation. 
To address this, one can tune the protrusion loss and apply post-processing to pull the virtual camera back toward the real motion when the virtual camera pose deviates from the real pose too much. 
Our experiments also show a discrepancy between the existing metrics and user preference. 
Closing this gap with more human perception studies and defining representative metrics will enable more effective learning-based solutions.

{\small
\bibliographystyle{ieee_fullname}
\bibliography{egbib}
}

\end{document}